\documentclass[twocolumn,final]{elsarticle}

\usepackage{ycviu}
\usepackage{framed,multirow}

\usepackage{amssymb}
\usepackage{latexsym}

\usepackage{url}
\usepackage{xcolor}
\definecolor{newcolor}{rgb}{.8,.349,.1}

\newcommand{\minisection}[1]{\vspace{0.04in} \noindent {\bf #1}\ \ }

\usepackage{amsmath}
\usepackage{subcaption}
\usepackage{booktabs}
\usepackage{multirow}
\usepackage{placeins}

\journal{Computer Vision and Image Understanding}

\begin{document}

\begin{frontmatter}

\title{Controlling biases and diversity in diverse image-to-image translation}

\author[1]{Yaxing \snm{Wang}\corref{cor1}} 
% \cortext[cor1]{Corresponding author: 
%   Tel.: +0-000-000-0000;  
%   fax: +0-000-000-0000;}
\cortext[cor1]{Corresponding author: 
   Tel.: +34-64444248;  
}
\ead{yaxing@cvc.uab.es}
\author[1]{Abel \snm{Gonzalez-Garcia}}
\author[1]{Luis \snm{Herranz}}
\author[1]{Joost \snm{van de Weijer}}

\address[1]{Computer Vision Center, Edifici O, Universidad Aut\'{o}noma de Barcelona, 08193, Bellaterra, Spain.}

\received{1 May 2013}
\finalform{10 May 2013
\availableonline{15 May 2013}
\communicated{S. Sarkar}
}

\begin{abstract}
The task of unpaired image-to-image translation is highly challenging due to the lack of explicit cross-domain pairs of instances.
We consider here diverse image translation (DIT), an even more challenging setting in which an image can have multiple plausible translations. 
This is normally achieved by explicitly disentangling content and style in the latent representation and sampling different styles codes while maintaining the image content.
Despite the success of current DIT models, they are prone to suffer from bias.
In this paper, we study the problem of bias in image-to-image translation. 
Biased datasets may add undesired changes (e.g. change gender or race in face images) to the output translations as a consequence of the particular underlying visual distribution in the target domain. 
In order to alleviate the effects of this problem we propose the use of semantic constraints that enforce the preservation of desired image properties. 
Our proposed model is a step towards unbiased diverse image-to-image translation (UDIT), and results in less unwanted changes in the translated images while still performing the wanted transformation.
Experiments on several heavily biased datasets show the effectiveness of the proposed techniques in different domains such as faces, objects, and scenes.
\end{abstract}

\begin{keyword}
\MSC 41A05\sep 41A10\sep 65D05\sep 65D17
\KWD Keyword1\sep Keyword2\sep Keyword3

%% MSC codes here, in the form: \MSC code \sep code
%% or \MSC[2008] code \sep code (2000 is the default)
\end{keyword}

\end{frontmatter}

%% main text
\section{Introduction}
\label{sec1-Introduction}
\begin{figure}[th]
\centering
\includegraphics[width=\columnwidth]{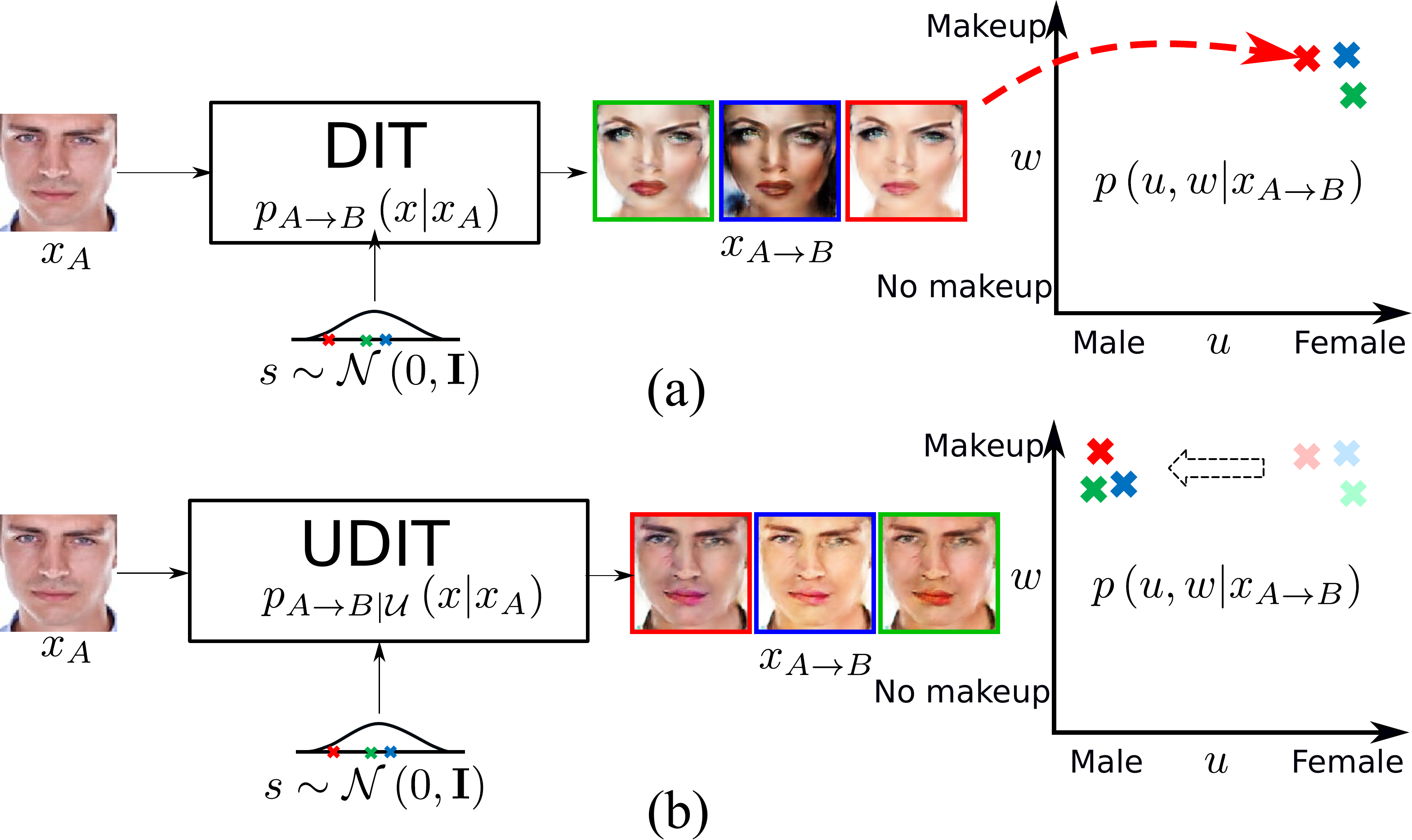}
\vspace{-4mm}
\caption{\small Diverse image-to-image translation in a very biased setting (domain A: mostly white males without makeup, domain B: white females with makeup): (a) biased translations,  (b) with semantic constraint to alleviate bias while keeping relevant diversity.\vspace{-3mm}}
\label{fig:motivation}
\end{figure}
%\linenumbers

%\linenumbers
Image-to-image translation (simply image translation hereinafter) is a powerful framework to apply complex data-driven transformations to images~\citep{gonzalez2018image,isola2017pix2pix,kim2017learning,lee2018diverse,yi2017dualgan,wang2018mix}. The transformation is determined by the data collected from the input and output domains, which can be arranged as explicit input-output instance pairs~\citep{isola2017pix2pix} or just the looser pairing at set level~\citep{kim2017learning,liu2017unsupervised,yi2017dualgan,zhu2017unpaired}, known as paired and unpaired image translation, respectively.

Early image translation methods were deterministic in the sense that same input image is always translated to the same output image. However, a single input image often can have multiple plausible output images, allowing for variations in color, texture, illumination, etc. Recent approaches allow for diversity\footnote{In some papers this is referred to as \textit{multimodal}, in the sense that the output distribution can have multiple modes.} in the output~\citep{huang2018multimodal,lee2018diverse,zhu2017toward} by formulating image translation as a mapping from an input image to a (conditional) output distribution (see Fig.~\ref{fig:motivation}a), where a particular output is sampled from that distribution. In practice, the sampling is performed in the latent representation that is the input of the generator, 
which is explicitly disentangled into content representation and style representation~\citep{lee2018diverse,zhu2017toward}.
Concretely, the style code is sampled to achieve diversity in the output while preserving the image content.

A concern with image translation models, and machine learning models in general, is that they capture the inherent biases in the training datasets. The problem of undesired bias in data is paramount in deep learning, raising concerns in multiple communities as automation and artificial intelligence become pervasive in their interaction with humans, such as systems involving analyzing face or person images, or communication in natural language. For example, it is known that most face recognition systems suffer from gender and racial bias~\citep{buolamwini2018gender}. Similar gender bias is observed in image captioning~\citep{hendricks2018women}. Here we focus on the kind of biases that may affect image translation systems.
Although bias is inherent to data collection, it is certainly possible to design better and more balanced datasets, or at least understand the related biases, their nature and try to incorporate tools to alleviate them~\citep{howard2017addressing,jiang2019identifying,zhao2018bias,zou2018ai}.

What particular visual and semantic properties of the input image are changed during the translation is determined by the internal and relative biases between the input and output training sets. These biases have significant impact on the diversity and potential unwanted changes, such as changing the gender, race or identity of a particular input face image. As an example we can consider the input domain \textit{faces without makeup} and the output domain \textit{faces with makeup}, so we expect that the image translator learns to add makeup to a face.  However, the input training set may be heavily biased towards males without makeup, and the output training set towards females with makeup\footnote{In addition to biases towards white and young people, we do not consider other specific biases in this example for the sake of simplicity.

}. With such biases, the translator learns to generate female faces with makeup even when the input is a male face (see Fig.~\ref{fig:motivation}a). While the change in the makeup attribute is desired, the change in identity and gender are not.

In this paper we propose to make the image translator counter undesired biases, by incorporating \textit{semantic constraints} that enforce minimizing the undesired changes (e.g. see Fig.~\ref{fig:motivation}b when constraining the identity, which implicitly constrains gender). These constraints are implemented as neural networks that extract relevant semantic features. 
Designing an adequate semantic constraint is often not trivial, and naive implementations may carry irrelevant information. 

This often leads to undesired side effects such as ineffective bias compensation and limiting the desired diversity in the output. 
Here we address these issues and propose an approach to design an effective semantic constraint that both alleviates bias and preserves desired diversity.

\section{Related Work}
Image-to-image translation has recently received exceptional attention due to its excellent results and its great versatility to solve multiple computer vision problems~\citep{bozorgtabar2019learn,huang2018multimodal,isola2017pix2pix,CMGGANs,liu2019exploiting,zhang2018synthetic,zhu2017unpaired,zhu2017toward}. 
Most image translation approaches employ conditional Generative Adversarial Networks (GANs)~\citep{goodfellow2014generative}, which consist of two networks, the generator and the discriminator, that compete against each other.
The generator attempts to generate samples that resemble the original input distribution, while the discriminator tries to detect whether samples are real or originate from the generator.
In the case of image translation, this generative process is conditioned on an input image.
The seminal work of~\citet{isola2017pix2pix}, pix2pix, was the first GAN-based image translation approach that was not specialized to a particular task.
In spite of the exceptional results on multiple translation tasks such as grayscale to color images or edges to real images, this approach is limited by the requirement of pairs of corresponding images in both domains, which are expensive to obtain and might not even exist for particular tasks. %Unpaired methods
Several methods~\citep{kim2017learning,liu2017unsupervised,taigman2017unsupervised,yi2017dualgan,zhu2017unpaired} have extended pix2pix to the unpaired setting by introducing a cycle consistency loss, which assumes that mapping an image to the target domain and then translating it back to the source should leave it unaltered.

% One-to-many / diversity 

%  \vspace{-2mm}
\minisection{Diversity in image-to-image translation.}
A limitation of the above image translation models is that they do not model the inherent diversity of the target distribution (e.g. same shoe can come in different colors).
For example, pix2pix~\citep{isola2017pix2pix} tries to generate diverse outputs by including noise alongside the input image, but this noise is largely ignored by the model and the output is effectively deterministic.
BicycleGAN~\citep{zhu2017toward} proposed  to overcome this limitation by adding the reconstruction of the latent input code as a side task, thus forcing the generator to take noise into account and create diverse outputs. BicycleGAN still requires paired data.
In the unpaired setting, several recent works~\citep{almahairi2018augmented,huang2018multimodal,lee2018diverse} address unpaired diverse image translation. 
Our approach falls into this category as it does not need paired data and it outputs diverse translations.
Our work is closest to MUNIT~\citep{huang2018multimodal}, which divides the latent space into a shared part across domains and a part specific to each domain.
However, these methods output too much diversity in some cases, which results in the undesired change of image content that should be preserved by the model (e.g.\ identity, race). 
Moreover, such changes are often determined by the underlying bias in the dataset, which MUNIT captures and amplifies during translation.

 \begin{figure}[t]
\centering
\includegraphics[width=\columnwidth]{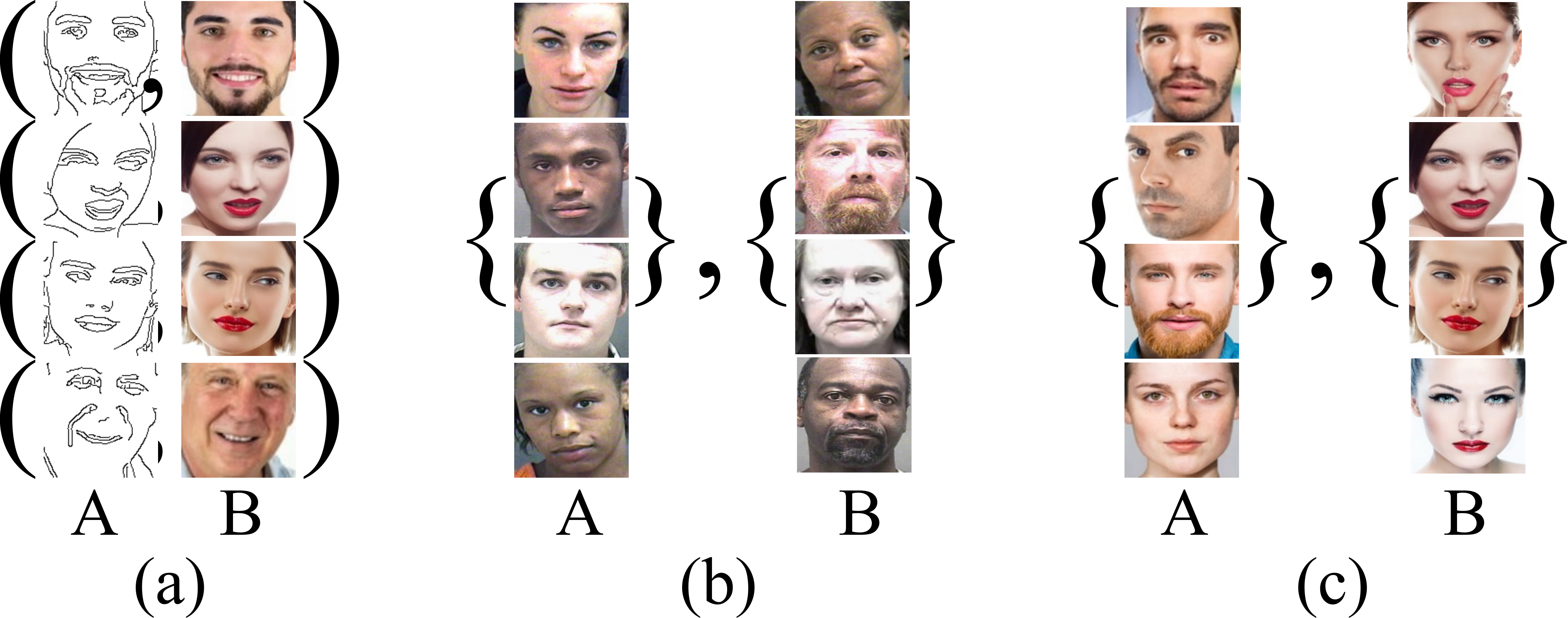}
\vspace{-4mm}
\caption{\small Examples of training sets for image translation: (a) paired edge-photo, (b) unpaired young-old (well-aligned biases), and (c) unpaired without-with makeup (misaligned in gender).\vspace{-3mm}}
\label{fig:biases_samples}
\end{figure}

% \vspace{-2mm}
\minisection{Disentangled representations.} While DIT methods explicitly disentangle content and style to enable diversity, other methods attempt to obtain disentangled representations to isolate different factors of variation in images~\citep{bengio2013representation}, which is beneficial for tasks such as cross-domain classification~\citep{bousmalis2017unsupervised,bousmalis2016domain,ganin2015unsupervised,liu2018detach} or retrieval~\citep{gonzalez2018image}.
In the context of generative models,~\citet{mathieu2016disentangling} combined a GAN with a Variational Autoencoder (VAE) to obtain an internal representation that is disentangled across specified (e.g.\ labels) and unspecified factors. 
InfoGAN~\citep{chen2016infogan} achieves some control over the variation factors in images by optimizing a lower bound on the mutual information between images and their representations. 
Some approaches impose a particular structure in the learned image manifold, either by representing each factor of variation as a different sub-manifold~\citep{reed2014learning} or by solving analogical relationships through representation arithmetic~\citep{reed2015deep}.
The work of~\citep{gonzalez2018image} achieves cross-domain disentanglement by separating the internal representation into a shared part across domains and domain-exclusive parts, which contain the factors of variation of each domain.
In our case we assume we do not have access to disentangled representations beyond content and style, and especially between wanted and unwanted changes.

% \vspace{-2mm}
\minisection{Bias in machine learning datasets.} 
Since machine learning is mostly fitting predictive models to data, the problem of biased training data is of great relevance. Dataset bias in general refers to the observation that models trained in one dataset may lead to poor generalization when evaluated on other datasets, due to the specific bias in each of them~\citep{torralba2011unbiased}. Bias is multifaceted, and datasets can be biased in many ways (e.g. illumination conditions, capture devices, class imbalance, scale~\citep{herranz2016scene}). Dataset bias can be addressed and improve cross-dataset generalization~\citep{fang2013unbiased,khosla2012undoing}. A related problem is domain adaptation~\citep{daume2009frustratingly,patel2015visual} where models trained on a source domain are adapted to a target domain, trying to overcome the difference in biases. Biased datasets lead to biased models, which have severe implications as data-driven artificial intelligence becomes pervasive. For instance, most commercial face recognition and image captioning systems exhibit gender and ethnicity biases~\citep{buolamwini2018gender, hendricks2018women}. 
Therefore, tackling bias is an increasingly important topic in machine learning~\citep{howard2017addressing,jiang2019identifying,zhao2018bias,zou2018ai}. Here we focus on the specific problem of understanding bias in image translation.

\vspace{-2mm}
\section{Diverse image translation}
\vspace{-2mm}
% \subsection{Unpaired diverse image translation}
\subsection{Definition and Setup}
Our goal is to translate samples from a source domain $A$ to a target domain $B$ in an unpaired setting, i.e.\ without corresponding images across domains.
Let $x_A \in X_A$ be a sample from the marginal distribution of images in the source domain, $p_A(x)$.
We want to obtain a translation $x_{A \to B}$ to $B$, sampled from a conditional distribution $p_{A\to B}(x|x_A)$ that approximates the true conditional $p_B(x|x_A)$.
The difficulty of this task resides in the impossibility to observe the joint distribution $p_{A,B}(x_A,x_B)$ in the unpaired setting, and the complexity of the conditional distribution $p_B(x|x_A)$, which is generally multi-modal. 
Simultaneously, we want to obtain the inverse translation $x_{B \to A}$.

Current unpaired diverse image translation methods~\citep{huang2018multimodal,lee2018diverse} use an encoder-decoder architecture, where the input image is first encoded into a latent code and then later decoded to generate the translated target image.
These methods resort to the assumption that part of the latent space, the \emph{content}, is shared by both domains, whereas the \emph{style} contains only the domain-specific characteristics.
Concretely, let us consider content encoders $E_i^c$ and style encoders $E_i^s$, where $i \in\{A,B\}$ indexes over domains.
Then, the latent representation of an input image $x_i$ can be decomposed into content $c_i=E_i^c(x_i)$ and style $s_i=E_i^s(x_i)$.
Given that style is purely domain-specific, we only need the particular content code $c_i$ for translation, combined with a randomly sampled style code $s'\sim \mathcal{N}(0,\mathbf{I})$, to generate the output image through the decoder $G_j$ as\  $x_{i\to j} = G_j(c_i,s')$.

Note that the decoders are deterministic functions that act as inverses of the encoders ($x_i = G_i(E_i^c(x_i),E_i^s(x_i))$, the stochasticity of the output translations is introduced through the sampling of the style code, which is the source of diversity on the generated translations (Fig.~\ref{fig:udit}a).

\subsection{Biases in diverse image translation}

\minisection{Wanted and unwanted properties.}
Images are complex and diverse in nature, reflected at many levels, such as visual appearance, structure and semantics. 
Therefore, the dataset bias is also complex and multifaceted, and it may be convenient to analyze separately specific biases depending on specific semantic properties.
Let $a\left(w,u\right)$ represent the relevant semantic properties associated with an image $x$ that are subject to change during translation, with $w$ being those we want to change (i.e.\ \textit{wanted}), and $u$ being those we do not want to be changed (i.e.\ \textit{unwanted}). We assume that they can be obtained via the mappings $w=g\left(x\right)$ and $u=h\left(x\right)$. For instance, in the example of Fig.~\ref{fig:motivation}, $w$ is makeup and $u$ is gender (for simplicity, but more generally $u$ could also include identity, race, etc.). The distributions of images of the source domain $i$ and the target domain $j$ induce the corresponding distributions of properties $p_i\left(w,u\vert x_i\right)$ and $p_j\left(w,u\vert x_j\right)$, respectively. 
% (see Fig.~\ref{fig:udit}b, where $i=A$ and $j=B$).

\begin{figure}[t]
    \centering
    \includegraphics[width=0.3\textwidth]{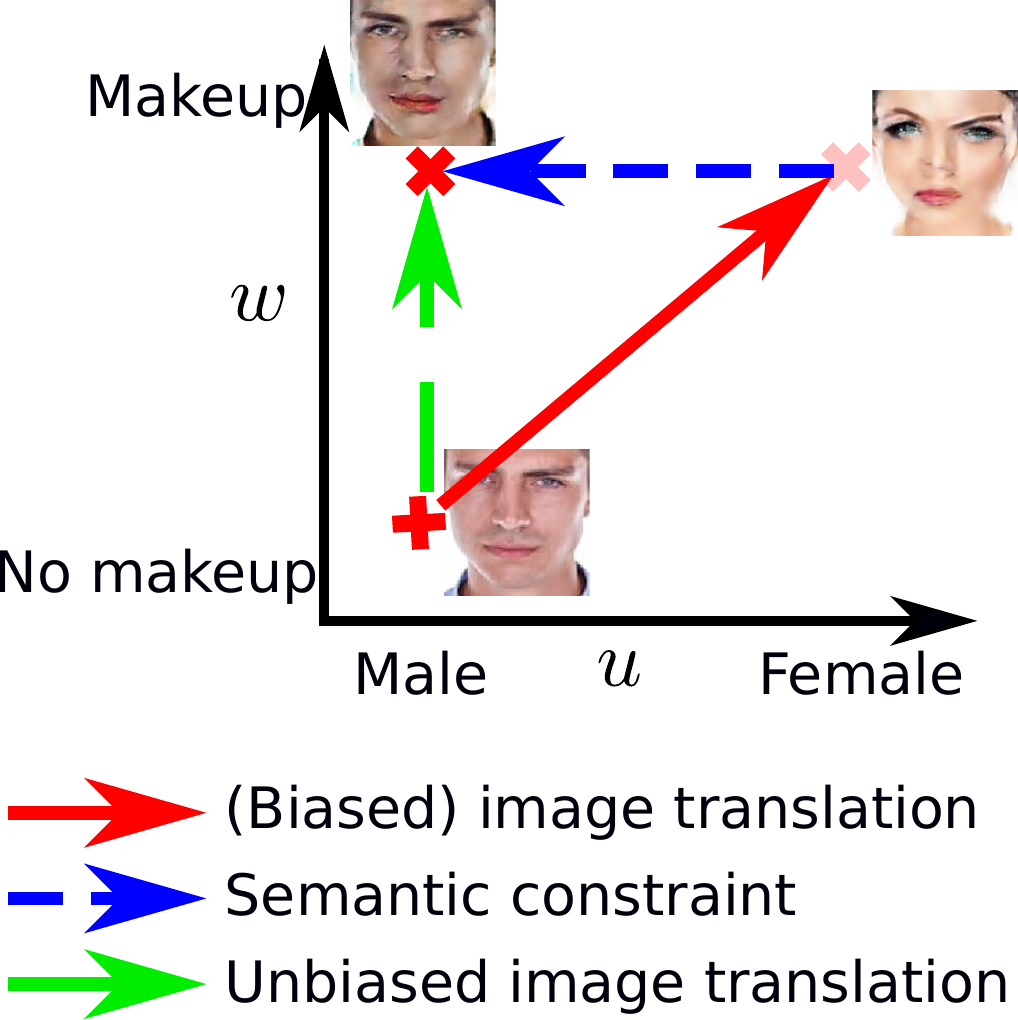}
   \caption{Geometric interpretation of the semantic constraint unbiasing the translation.
%   \YX{What is the black and red arrowes in (b)? Also the solid line and dotted line in (b)?
}\vspace{-5mm}
   \label{fig:property_space}
\end{figure}

% \subsection{Translations in the space of properties}
\minisection{Translations in the space of properties.}
During training, the image translator learns the mapping between both domains, and consequently what properties to modify. An input image $x_i$ has the properties $w_i=g\left(x_i\right)$ and $u_i=g\left(x_i\right)$, and the corresponding translation $x_{i\to j}$ will have $w_{i\to j}=g\left(x_{i\to j}\right)$ and $u_{i\to j}=h\left(x_{i\to j}\right)$. The image translation is \textit{successful} if and $w_{i\to j}\neq w_i$ is effectively the wanted property of the target domain. Similarly, a translation is \textit{unbiased} when $u_{i\to j}=u_i$. In general, DIT results in biased translations when $u_{i\to j}\neq u_i$ (see Fig.~\ref{fig:property_space}), which stems from the original bias in the training dataset.
% \alertJW{should this be figure 4b?}\YX{You mean Fig.3a?, I refer to the submitted paper, it is Fig.3a}

\section{Unbiased diverse image translation}
\subsection{Unbiasing the generated images}
For simplicity, let us consider the paired image translation case where a ground truth translation $x_j$ is available for each $x_i$, with the corresponding properties $\left(w_j,u_j\right)=g\left(x_j\right)$. In order to learn a successful and unbiased translation we would like to enforce the constraints $w_{i\to j}=w_j$ and $u_{i\to j}=u_i$, respectively.

However, we focus on the the more complex case of diverse image translation, where the output is stochastic, i.e. a distribution rather than a single image. In this case the constraints may not be enforced at the sample level but at the distribution level. In the case of $u$ we aim at enforcing
\begin{gather}
\vspace{-1mm}
    u_{i\to j}=h\left(x_{i\to j}\right)=h\left(x_i\right)=u_i \label{eq:unwanted}\\
    \forall x_{i\to j}\sim p_{i\to j}\left(x\vert x_i\right), \forall x_i\sim p_i\left(x\right) \nonumber
\vspace{-1mm}
\end{gather}
which ensures that the unwanted properties remain unchanged throughout the translation.
Similarly, %we aim at enforcing
\begin{gather}
\vspace{-1mm}
    w_{i\to j}=g\left(x_{i\to j}\right)=g\left(x_j\right)=w_j \label{eq:wanted}\\
    \forall x_{i\to j}\sim p_{i\to j}\left(x\vert x_i\right), 
    \forall x_j\sim p_j\left(x\vert x_i\right), 
    \forall x_i\sim p_i\left(x\right) \nonumber
\vspace{-1mm}
\end{gather}
which ensures that the wanted properties change properly, according to the desired translation. Note that for convenience we assume that the true conditional distribution of the translation $p_j\left(x\vert x_i\right)$ is known.

In this way, the biases in the distribution of generated images would be aligned properly, achieving our goal of removing unwanted biases in the translation (see Fig.~\ref{fig:property_space}). In the previous example we would like the translated images to preserve the statistics of gender distribution of $A$ while adapting to the statistics of makeup distribution of $B$. Similarly in the direction from $B$ to $A$.

Note that the different settings in image translation implicitly or explicitly enforce this sort of alignments via pairing or the design of the dataset. For instance, Fig.~\ref{fig:biases_samples}a shows an example of a dataset for paired translation, where the instance-level pairing already prevents unwanted gender bias (50\% males and females). Gender bias can also be prevented in unpaired translation by designing well-balanced and statistically aligned training sets for domains A and B (see Fig.~\ref{fig:biases_samples}b). However, Fig.~\ref{fig:biases_samples}c shows a dataset clearly biased and misaligned on gender. In this case, it is desirable that the model can be forced to correct this unwanted misalignment, to prevent biased translations.

% The above formulation regularizes the DIT model to prevent unwanted changes while allowing the change of some specified wanted properties.  

In practice, directly enforcing constraints~(\ref{eq:unwanted}) and (\ref{eq:wanted}) is not possible since $w$ and $u$ are not disentangled in our setting. Besides, we do not have access to $p_j\left(x\vert x_i\right)$. 

For this reason we propose to implement~\eqref{eq:unwanted} via the addition of a semantic regularization constraint that enforces the preservation of $u$ properties during translation, while constraint (\ref{eq:wanted}) is indirectly enforced via the image translation loss.
A bad implementation of the semantic constraint can hamper the effectiveness of image translation in practice (e.g.\ limiting diversity), so the appropriate design of the semantic constraint and its implementation is related to both constraints.

\subsection{Semantic regularization constraint}
% \vspace{-2mm}
Here we propose an Unbiased DIT model (UDIT) that enforces constraint~\eqref{eq:unwanted} via a \emph{semantic extractor} $h$ that estimates the representative semantic properties we want to preserve in the image as $u_i=h\left(x_i\right)$.

\begin{figure}[t]
    \centering
    \includegraphics[width=0.8\columnwidth]{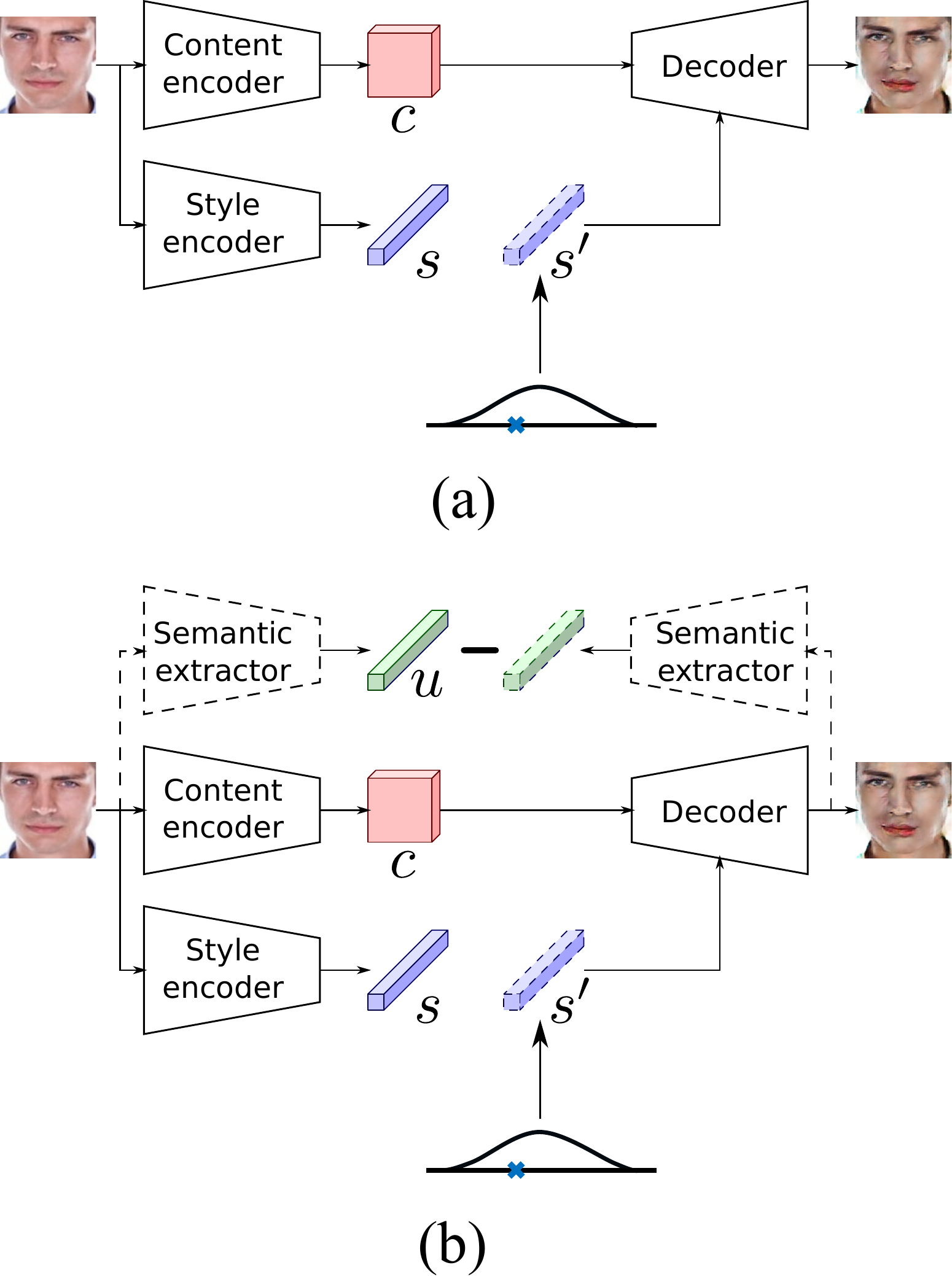}
    
   \caption{Diverse image-to-image translation (DIT): (a) biased, (b) unbiased (i.e. UDIT) via a semantic constraint implemented with a semantic extractor.
}\vspace{-3mm}
   \label{fig:udit}
\end{figure}

Constraint~\eqref{eq:wanted} on the wanted changes is implicitly enforced by the DIT model, including the unpaired setting. 
Fig.~\ref{fig:udit}b illustrates how a proper semantic constraint regularizes the initial DIT model to alleviate the unwanted bias.

In particular, we include a \emph{semantic constraint loss} 
\vspace{-1mm}
\begin{equation}
\label{eq:sem_const}
\vspace{-2mm}
\mathcal{L}^{u_i}_{\mathcal{U}} = \mathbb{E}_{x_i \sim p_i(x) }\big[|| u_{i\to j} - u_{i} ||\big],
\end{equation}
where $\mathcal{U}$ represents the semantic properties we want to keep unchanged throughout the translation. By including $\mathcal{L}^{u_i}_{\mathcal{U}}$ in our training objective (sec.~\ref{sec:full_model}), we are effectively conditioning the output conditional distribution to $\mathcal{U}$, i.e.\ $p_{i \to j | \mathcal{U}}(x|x_i)$, and hence alleviating the unwanted bias in the output samples $x_{i \to j} \sim p_{i \to j | \mathcal{U}}(x|x_i)$, when $\mathcal{U}$ is properly designed. Fig.~\ref{fig:udit}b shows the architecture of this UDIT. Note how this constraint is only enforced during training, we do not use $u_i$ during translation at inference time.

\vspace{-2mm}
\section{UDIT implementation} 
\vspace{-2mm}
\label{sec:udit_implementation}

\subsection{Semantic extractor}
\label{sec:semantic_extractor}
% \vspace{-2mm}
Crucial for the success of our method is the proper design of the semantic extractor $h\left(x\right)$, which in general will be implemented as a neural network. We must guarantee that the extracted feature contains enough relevant information regarding the specific semantic property that we want to preserve (i.e. captures $u$ properly). On the other hand, we want to prevent it from containing additional information that could potentially introduce undesired side effect such as limiting the translation ability of the model or the diversity on the output. 
We now develop a procedure to design effective semantic extractors that satisfy both requirements.

% \vspace{-2mm}
\minisection{Capturing the semantic property.}
As feature extractors, we consider convolutional neural networks (CNNs) implementing classification tasks related with $u$ (e.g. gender classification), which we train on a suitable external dataset. The CNN may also be initialized with models pretrained in large datasets (e.g. ImageNet~\citep{russakovsky2015imagenet}, DeepFace~\citep{taigman2014deepface}). In principle we are interested in a suitable intermediate feature that captures $u$ well. In particular, the convolutional features that are input into the first fully connected layer are often good candidates, as they contain semantically meaningful information while still being spatially localized.

\minisection{Reducing undesired information.}
Deep features from generic feature extractors such as models trained in ImageNet capture rich and varied properties in a relatively high dimensional feature. This can be harmful in our case, since they can also capture properties unrelated with $u$. The classifier can learn to ignore them and still solve the task, but they remain as noise in the semantic feature, being enforced through the constraint and therefore limiting the flexibility of the image translator to generate the wanted change and diversity. In order to address this problem, we propose to add an additional convolutional layer with a kernel $1\times1\times D$ with the purpose of reducing the dimensionality of the feature. 
We experimentally find the minimum value of $D$ that keeps a satisfactory accuracy. 
The output of this additional layer is used as semantic feature.

In summary, the designed features will ideally contain the right amount of information relevant for the task, and no irrelevant information that could interfere with the wanted translation.

% \section{Unbiasing MUNIT}
\vspace{-2mm}
\subsection{Full model}
% \vspace{-0mm}
\label{sec:full_model}
The proposed unbiasing methodology is generic enough to be applicable in most image-to-image translation methods. 
The UDIT models in our experiments are based on MUNIT~\citep{huang2018multimodal} extended with particular semantic constraints.
The model is composed of within-domain autoencoders and cross-domains translators with reconstruction of translated features.
We also consider a variant that uses pooling indices as side information~\citep{badrinarayanan2017segnet}.
 
In the following, we detail the remaining losses and present the full model.

\minisection{Adversarial loss.}The translator attempts to generate realistic images that fool the discriminator $D_j$, whose task is to distinguish fake images from real images.
The discriminator is trained adversarially with 

\vspace{-2mm}
\begin{equation}
\label{eq:adver_loss}
\begin{aligned}
\vspace{-2mm}
    \mathcal{L}^{x_j}_{GAN}&  =  \frac{1}{2}\mathbb{E}_{x_i \sim p_i(x),s'\sim \mathcal{N}(0,\textbf{I})}\big[ (
    D_j(G_j(c_i,s'))^2\big] \\ 
     & + \mathbb{E}_{x_j \sim p_j(x)}\big[(D_j(x_j)-1)^2\big].
\end{aligned}
\end{equation}

\minisection{Reconstruction loss.} The autoencoders ensure that the model is able to reconstruct the input image through the image reconstruction loss
\vspace{-2mm}
\begin{equation}
\label{eq:recon_img}
    \vspace{-2mm}
    \mathcal{L}^{x_i}_{recon} =\mathbb{E}_{x_i \sim p_i(x) }\big[ ||G_i(c_i,s_i)- x_i||_1\big]. 
\end{equation}
Moreover, the translated image is further encoded in both content and style, and the following feature reconstruction losses are applied 
\vspace{-2mm}
\begin{equation}
\label{eq:recon_con}
\vspace{-2mm}
    \mathcal{L}^{c_i}_{recon} =\mathbb{E}_{x_i \sim p_i(c),s'\sim \mathcal{N}(0,\textbf{I})}\big[ ||E_j^c(G_j(c_i,s')) - c_i||\big],
\end{equation}
\begin{equation}
\label{eq:recon_sty}
    \mathcal{L}^{s_i}_{recon} =\mathbb{E}_{x_i \sim p_i(c),s'\sim \mathcal{N}(0,\textbf{I})}\big[ ||E_j^s(G_j(c_i,s')) - s'||\big].
\end{equation}
The loss on $c_i$ enforces the preservation of the content code across domains, whereas the loss on the style encourages diversity on the outputs. 

%
% Let $(c_i,k_i,s_i) = (E_i^c(x_i),E_i^k(x_i),E_i^s(x_i))$ be the encoding of an input image $x_i$ into content, side information
The loss used to trained UDIT follows MUNIT's loss  combined with the semantic constraint loss~\eqref{eq:sem_const} as follows

\vspace{-2mm}
\begin{equation}
\label{eq:full_loss}
    \begin{aligned}
    \vspace{-2mm}
    \mathcal{L} = &  \mathcal{L}^{x_A}_{GAN} + \mathcal{L}^{x_B}_{GAN} + \lambda_x (\mathcal{L}^{x_A}_{recon} + \mathcal{L}^{x_B}_{recon}) \\
    & \lambda_c ( \mathcal{L}^{c_A}_{recon}  + \mathcal{L}^{c_B}_{recon} ) 
    + \lambda_s ( \mathcal{L}^{s_A}_{recon}  + \mathcal{L}^{s_B}_{recon} ) \\
    & \lambda_\mathcal{U} (\mathcal{L}^{u_A}_{\mathcal{U}} + \mathcal{L}^{u_B}_{\mathcal{U}}),
    \end{aligned}
\end{equation}
where the $\lambda_x,\lambda_c,\lambda_s,\lambda_\mathcal{U}$ weights control the influence of each individual loss in the final objective. When $\lambda_\mathcal{U}=0$ we recover the baseline MUNIT model. We detail the network architectures in the Appendix.

\vspace{-2mm}
\section{Experimental results}
\vspace{-2mm}

\subsection{Datasets}

We conduct experiments on four datasets that suffer from different types of biases.

\minisection{Biased makeup} is our heavily biased dataset, where the female gender predominates in the target domain. We collected images of people with and without makeup from the web.
We retrieved 1,400 images of women with makeup by searching for ``woman makeup face" and manually verifying them.
For the no-makeup domain, we selected another 1400 images with 95\% males faces 
and 5\% female faces, so we purposely biased this domain towards males. All images were preprocessed by cropping the face, localized by a face detector.

\minisection{MORPH~\cite{ricanek2006morph}} is also a face dataset for age translation (young $\leftrightarrow$ old) with both ethnicity and gender biases. It contains 55,134 images of 13,000 subjects, and each image is annotated with gender, ethnicity, and age.
There are five ethnic groups represented in the dataset: Black (African ancestry), White (European ancestry), Hispanic, Asian, and `Other', which we discarded.

 MORPH is a face image dataset for adult age progression, where the images depict people of different ages at different points in time, spanning up to 30 years for some subjects. MORPH is heavily biased towards men ($>$85\%), and towards individuals with African ancestry ($>$78\%), followed by European ($\approx$17\%), Hispanic ($\approx$3.5\%) and Asian ($<$0.3\%) ancestries.
We perform experiments using the identity constraint (sec.~\ref{sec:makeup}) with the purpose of preserving both gender and ethnicity.

\minisection{Cityscapes~\cite{Cordts2016Cityscapes}$\to$Synthia~\cite{ros2016synthia}} contains real and synthesized urban scenes that are biased towards a particular time of the day (day/night). Cityscapes~\citep{Cordts2016Cityscapes} contains real street scenes captured from a moving vehicle during day-time (3000 images).
Synthia~\citep{ros2016synthia}, instead, is synthetically generated by a simulated car driving in a virtual world, both during day-time and night-time.  We artificially bias the day-time/night-time distribution of Sytnhia by selecting 3000 images captured during night and only 300 images during day.

\minisection{Biased handbags~\cite{zhu2017unpaired}} contains images of handbags with two defining attributes: color (\textit{red/black}) and texture (\textit{flat/textured}). We select red and black as possible colors.
Texture is also a binary attribute indicating the absence or presence of a non-flat texture on the handbags, i.e.\ flat or textured.

We create two datasets by selecting samples from the photo images of the handbags dataset used by~\citep{isola2017pix2pix,huang2018multimodal}.
The input domain only contains one mode (e.g.\ flat black handbags for Handbags-color), while the target domain contains two modes but is heavily biased towards one, e.g.\ 1000 textured red and 100 flat red. 

We note that we require the textured handbags to only have the right color (e.g.\ no stripes of another color), which limits the attribute to subtle variations mostly given by differences in the material.

Tables~\ref{tab:datasets_train_main_text} and~\ref{tab:datasets_test_main_text} specify the exact number of images used in our biased datasets for training and testing, respectively. Table~\ref{tab:classifier_main_text} reports the setting to train the metric network. 

Note for the biased makeup dataset, the used gender classifier is externally trained on Adience dataset~\cite{levi2015age}.

\begin{table}[t]
    \centering
    \resizebox{\columnwidth}{!}{
    \begin{tabular}{lcc}
    \toprule
         Experiment & Domain A & Domain B  \\
         \midrule
         Biased makeup & 1400 f-makeup & 1330 m-nomakeup, 70 f-nomakeup \\ 
         MORPH & 10000 m-y, 1000 f-y & 10000 m-o, 1000 f-o\\
         Cityscapes-Synthia & 3000 citys-day & 3000 syn-night, 300 syn-day\\
         Handbags-color &  755 flat-black &  1000 txt-red, 100 flat-red\\  
         Handbags-texture & 1256 flat-red & 1100 txt-black, 100 txt-red\\
         \bottomrule
    \end{tabular}}
    \caption{Details of datasets used for \emph{training} the image translation models. Abbreviations used: f=female, m=male, y=young, MORPHo=old, citys=cityscapes, syn=synthia, txt=textured.}
    \label{tab:datasets_train_main_text}
\end{table}

\begin{table}[t]
    \centering
    \resizebox{\columnwidth}{!}{
    \begin{tabular}{lcc}
    \toprule
         Experiment & Domain A & Domain B  \\
         \midrule
         Biased makeup & 100 f-makeup & 100 m-nomakeup \\ 
         MORPH &  200 m-y, 200 f-y & 200 m-o, 200 f-o\\
         Cityscapes-Synthia & 475 citys-day & -\\
         Handbags-color  & 100 flat-black & - \\
         Handbags-texture & 100 flat-red  & - \\
         \bottomrule
    \end{tabular}}
    \caption{Details of datasets used for \emph{testing} the image translation models. Abbreviations used: f=female, m=male, y=young, o=old, citys=cityscapes, syn=synthia, txt=textured.}\vspace{-3mm}
    \label{tab:datasets_test_main_text}
\end{table}
\begin{table}[h]
    \centering
    \resizebox{\columnwidth}{!}{
    \begin{tabular}{lcc}
    \toprule
         Experiment & Domain A & Domain B  \\
         \midrule
         MORPH-gender &  2000 m-y, 2000 m-o & 2000 f-y, 2000 f-o\\
         MORPH-ethnicity &  1200 afri-y, 1200 afri-o & 1200 euro-y, 1200 euro-o\\
         Cityscapes-Synthia & 3000 BDD-day, 3000 syn-day &3000 BDD-night,  3000 syn-night\\
         Handbags-MORPHcolor  & 500 flat-red, 500 flat-black & 500 txt-red, 500 txt-black \\
         Handbags-texture & 500 flat-red, 500 txt-red & 500 flat-black, 500 txt-black \\
         \bottomrule
    \end{tabular}}
    \caption{Details of datasets used training the classifier to evaluate quantitatively the results. Abbreviations used: f=female, m=male, y=young, o=old, afri=african, euro=european, BDD=BDD100K, syn=synthia, txt=textured. Note the used subsets are disjoint with the ones used to perform image translation.}\vspace{-3mm}
    \label{tab:classifier_main_text}
\end{table}

\begin{figure*}[t]
    \centering
    \includegraphics[width=\textwidth]{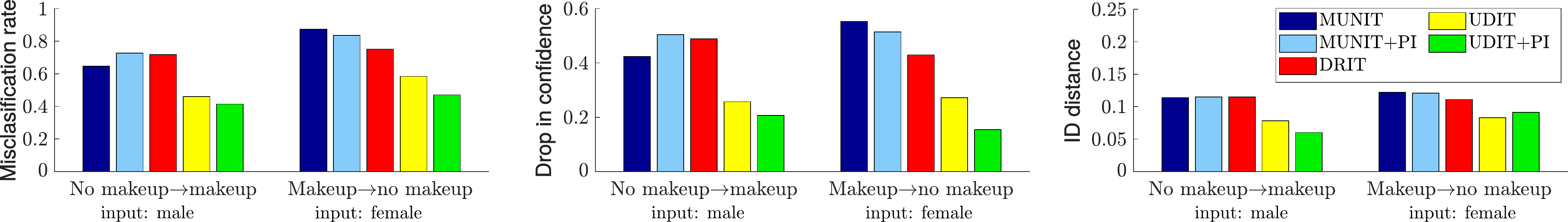}
    \caption{\small Robustness to bias on Biased makeup: (left) misclassification rate,  (middle) drop in confidence, (right) ID distance.\vspace{-3mm} %Results on both directions for the minority gender of each corresponding input domain.
    }
    \label{fig:makeup_scores}
\end{figure*}

\subsection{Baselines and variants}

We compare our method with the following approaches:

\minisection{MUNIT~\cite{huang2018multimodal}} disentangles the latent distribution into the content space which is shared between two domains,  and the style space which is domain-specific and aligned with a Gaussian distribution.  
At test time, MUNIT takes as an input the source image and different style codes to achieve diverse outputs.

\minisection{DRIT~\citep{lee2018diverse}} similarly explores the distribution of latent representation. Different from MUNIT by means of adaptive instance normalization to control diversity, DRIT directly insert noise into latent feature to achieve diverse output.  

We compare the previous baselines with different configurations of the proposed UNIT approach. In particular we study variants with and without Pooling Index(PI).

\subsection{Robustness to specific biases.}

Evaluating the generated images is challenge~\citep{borji2019pros}, here we introduce a new method to  measure whether translating an image across domains with misaligned biases changes particular properties of the image. 
For simplicity, we explain here these evaluation measures for the Biased makeup dataset (other datasets are similar). 
In particular, we want to evaluate whether applying or removing makeup on subjects changes their perceived gender.
In order to do this, we train a gender classifier $f\left(x\right)$ and evaluate the gender prediction over the translated image, i.e. $ f\left(x_{i \to j}\right)$.
Since we have the ground-truth label for the original image, we can determine whether gender has been changed with respect to the original image. 
We call this measure \emph{misclassification rate}.
The problem with this measure is that the classifier might output erroneous estimates in the first place for some challenging cases. 

For this reason, we also compute the \emph{drop in confidence} of the classifier during translation as $\delta\left(x_i\right) = f\left(x_i\right) - f\left(x_{i \to j}\right)$.

This score will indicate the effect of the translation on the classifier estimation of the correct label, somewhat accounting for the classifier's failure cases.

We can use the above measures with general properties such as gender or race. 
However, our face experiments also include a setting in which we want to preserve the \emph{identity} of the input.
Evaluating changes in identity is more complex since the set of categories is specific to the dataset.

In this case, we measure the change in identity by directly computing the distance between identity features given an off-the-shelf face recognition network~\citep{parkhi15deep}.
We call this measure \emph{ID distance} and only compute it for the face datasets.

\minisection{Diversity.} 

Several image translation approaches~\citep{zhu2017toward,huang2018multimodal,lee2018diverse} measure the diversity of the outputs by using the perceptual similarity metric LPIPS~\citep{zhang2018perceptual}, which is based on differences between deep features 

We follow the protocol introduced in~\citep{zhu2017toward} and average the LPIPS distance between 19 random pairs of outputs for 100 different input images.

\begin{figure}[t]
    \centering
    % \begin{subfigure}[b]{0.5\textwidth}
    \includegraphics[width=\columnwidth]{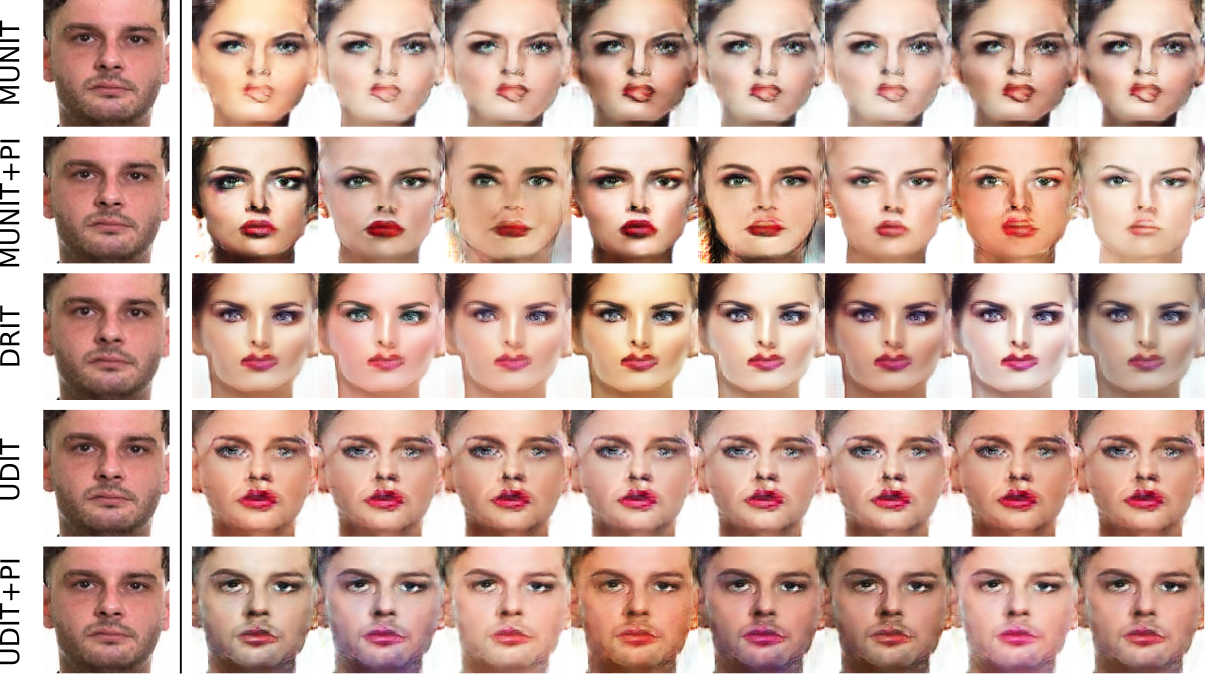}
    % \end{subfigure}%
    % ~ 
    % \begin{subfigure}[b]{0.5\textwidth}
    %     \centering
    %     \includegraphics[width=\textwidth]{images/example_female_to_nomakeup.pdf}
    %     \caption{\textit{makeup} to \textit{no makeup} (input: female)}
    % \end{subfigure}
    \caption{\small Example translations for Biased makeup when applying makeup to a male. UDIT uses identity as semantic constraint.
    \vspace{-5mm}
    }
    \label{fig:makeup_images}
\end{figure}

\begin{table}[t]
\begin{center}
\resizebox{0.9\columnwidth}{!}{
\begin{tabular}{lcccccc}
\toprule
Input & Direction & MUNIT & +PI & DRIT   & UDIT   & UDIT+PI  \\
\midrule
M     & Makeup    & 0.268 & 0.267 & 0.263 & 0.192 & 0.151   \\
F     & Makeup    & 0.212 & 0.199 & 0.193  & 0.154 & 0.133   \\
F     & Demakeup  & 0.297 & 0.293 & 0.253  & 0.208 & 0.203   \\ \bottomrule
\end{tabular}
}\vspace{-6mm}
\caption{\small LPIPS distance on Biased makeup. \vspace{-8mm}}
\label{tab:lpips_makeup}
\end{center}
\end{table}
\vspace{-1mm}
\subsection{Biased makeup dataset}
\vspace{-1mm}
\label{sec:makeup}
% Man, no makeup = 1200, Woman-makeup=1200 to train, another 200 for test

\minisection{Semantic constraint.}
In this dataset, we focus on the misalignment between biases at two levels: gender and identity. Preserving identity is a more restrictive constrain than preserving gender, and implicitly also preserves it.
For this reason, we use a semantic constraint based on identity (ID).
We consider an off-the shelf network for face recognition~\citep{parkhi15deep} and select its highest level convolutional features as semantic feature.
The model has been trained with VGG-Face~\citep{parkhi15deep}, which contains over 2000 different identities.

\begin{figure*}[t]
        \centering
          \includegraphics[width=\textwidth]{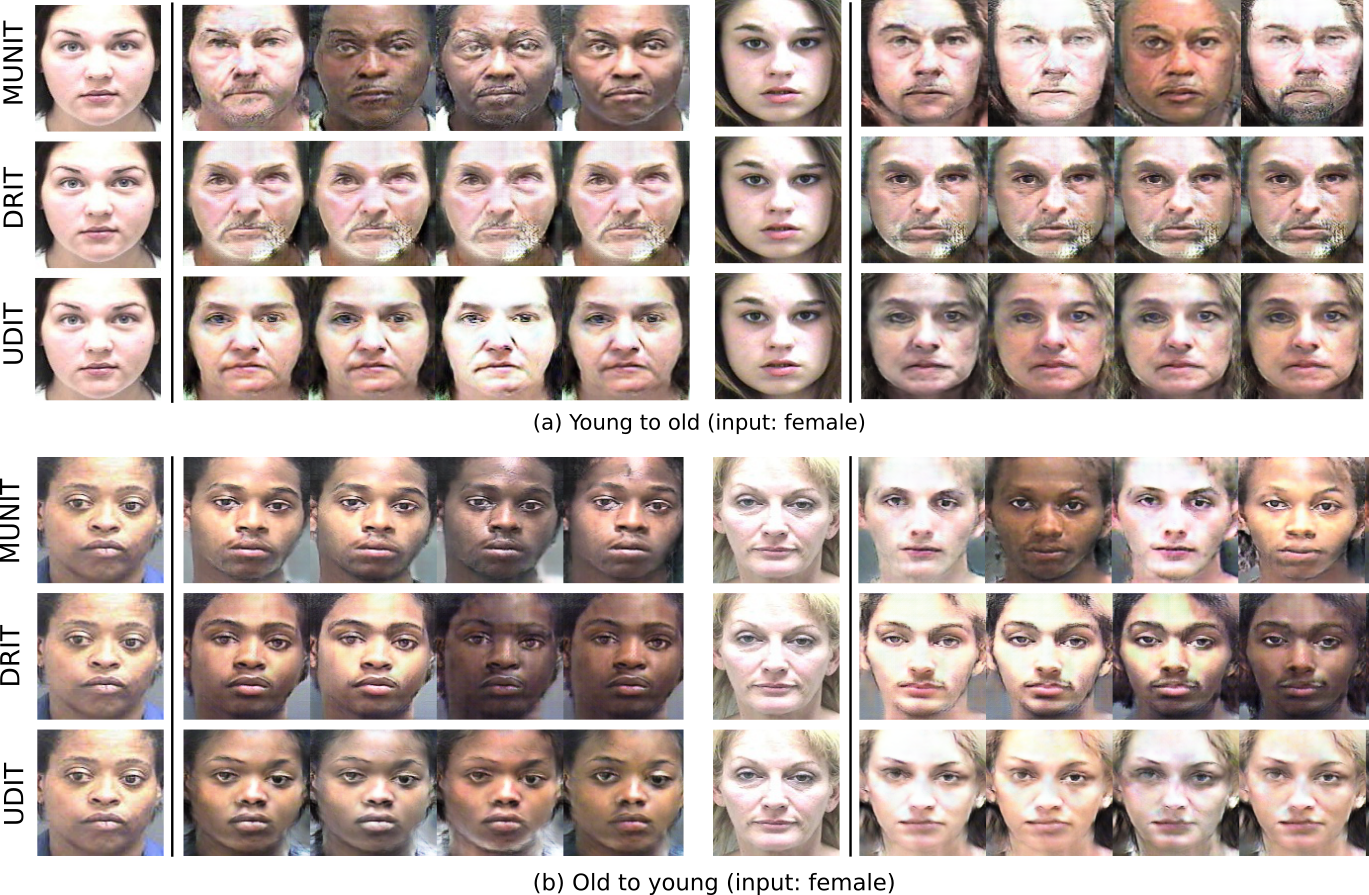}
    % \vspace{-6mm}
     \caption{\small Example translations on MORPH by biased DIT methods (MUNIT/DRIT) and our UDIT with semantic constraint on identity.} %\LH{No DIT=MUNIT+PI?}}
    \label{fig:morph_images}
\end{figure*}

\minisection{Qualitative evaluation.} 

Fig.~\ref{fig:makeup_images} compares image translations obtained with MUNIT~\citep{huang2018multimodal}, MUNIT with pooling indices (PI), DRIT~\citep{lee2018diverse}, and two variants of our model.
The basic UDIT variant only uses a semantic constraint on ID, whereas UDIT+PI uses also pooling indices.
We can observe that both MUNIT and DRIT change the gender (i.e. undesired change) when applying the desired translation (i.e. adding makeup).

This demonstrates the heavy influence of bias misalignment on DIT methods, which leads to the inevitable change of unwanted properties.
Moreover, the generated images lack realism and quality, resembling cartoonish versions of human faces.
Adding PI to MUNIT does not seem to bring any noticeable benefit. 

Instead, our UDIT model trained with the ID semantic constraint is very effective to prevent both unwanted gender and identity changes, as show in the figure. 
Furthermore, the incorporation of pooling indices results in an even more successful change on wanted properties (e.g.\ adding makeup to males), while generating images of high quality and realism.

\minisection{Robustness to unwanted changes.} Fig.~\ref{fig:makeup_scores} shows quantitative results of the three metrics evaluated on the different methods and both directions.
We only evaluate over the gender that is underrepresented in the target domain.
These results confirm the trends observed qualitatively in Fig.~\ref{fig:makeup_images}.
DIT baselines perform poorly at maintaining gender and identity, including MUNIT with PI.
Interestingly, the identity constraint clearly enhances the preservation of both wanted properties, as reflected by the substantial drop on all three robustness measures.
Moreover, UDIT+PI further increases robustness to bias.
This could be due to the improved quality of the output images with respect to the input, which leads to more reliable classifier predictions and pushes together the identity features. 
In the remainder of this paper we only employ the UDIT+PI variant and refer to it simply as UDIT, unless stated otherwise.

\begin{figure*}
    \centering
        \includegraphics[width=\textwidth]{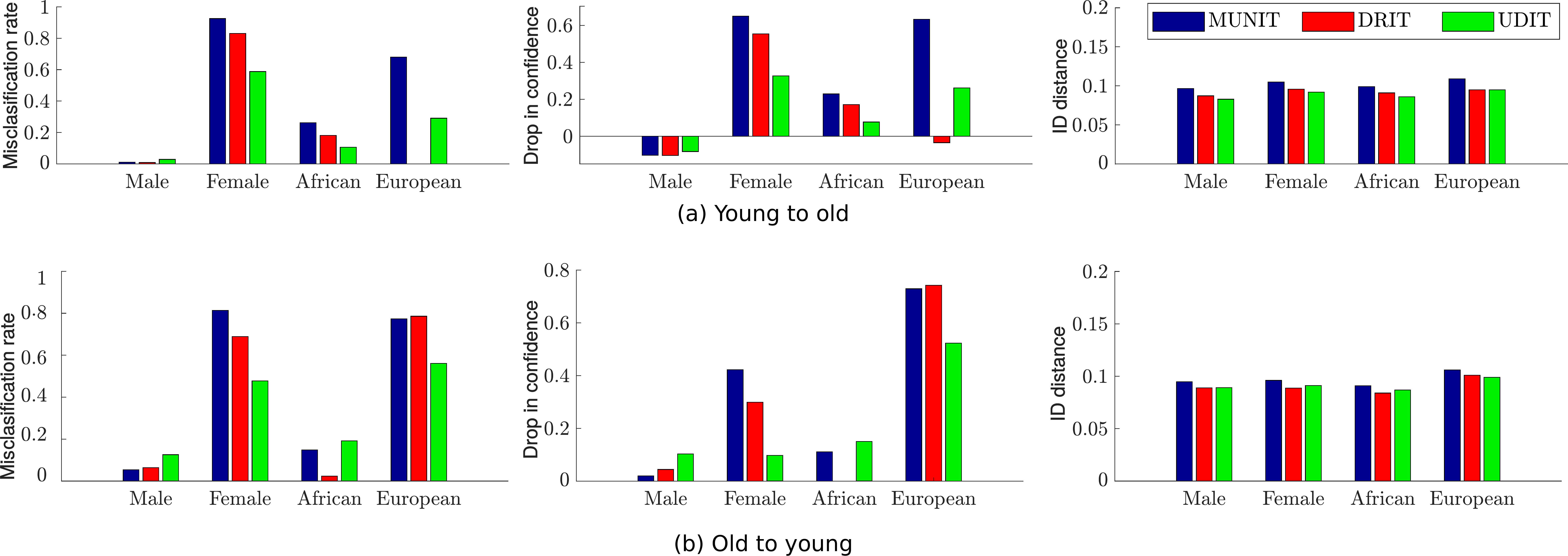}
        \vspace{-6mm}
    \caption{\small Robustness to bias on MORPH: (a)\emph{young} to \emph{old} and (b)\emph{old} to \emph{young}: (left) misclassification rate,  (middle) drop in confidence, and (right) ID distance.\vspace{-3mm} } 
    \label{fig:morph_scores}
\end{figure*}

\begin{figure}[t]
    \centering
    \includegraphics[width=\columnwidth]{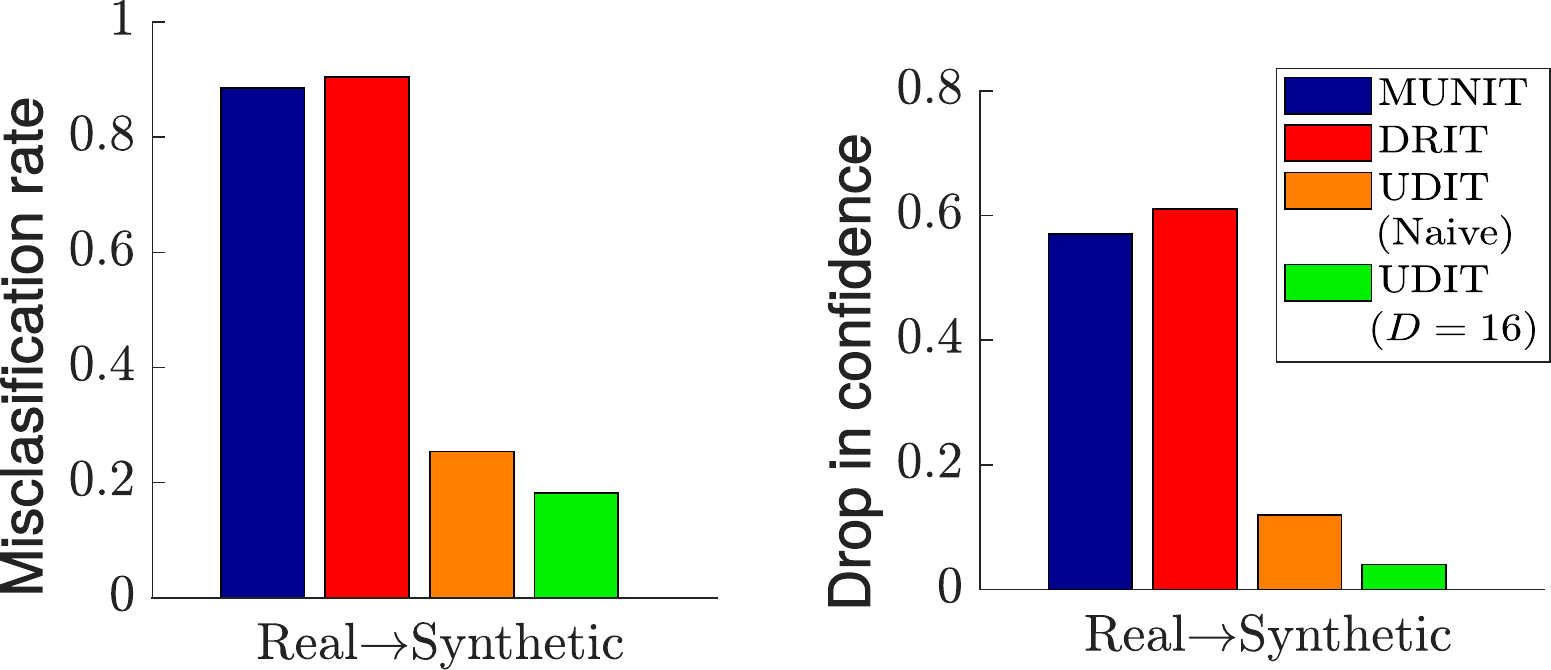}
    \vspace{-4mm}
    \caption{\small  Robustness to bias in terms of misclassification rate and drop in confidence .\vspace{-6mm}}
    \label{fig:cityscape_scores}
\end{figure}

\minisection{Diversity.} Table~\ref{tab:lpips_makeup} shows the LPIPS distance of the different evaluated methods.

UDIT models seem to be notably decreasing the LPIPS distance in comparison to MUNIT and DRIT. 
This makes sense since the identity constraint not only prevents unwanted bias, but it also constrains the diversity in those directions that compromise the preservation of identity. 

In this case, LPIPS distance may not be able to capture the more subtle variations that conform the diversity that should be expected in that setting. 
For example, the values for both UDIT variants are significantly lower than those of MUNIT or MUNIT+PI, but the examples in Fig.~\ref{fig:makeup_images} show that it is able to generate very diverse images, within the narrow space that allows preserving gender and identity (e.g.\ lip color, skin tone and shading, beard thickness). 

\setlength{\tabcolsep}{1.5pt}

\subsection{MORPH}

\begin{table}[t]
    \centering
    \resizebox{0.9\columnwidth}{!}{
    \begin{tabular}{lccccccc}
    \toprule
    $D$ & 2 & 8 & 16 & 32 & 64 & 128 & 256\\
    % \midrule
    % Faces-makeup & 94.4 & 94.5 & \textbf{95.9} & 96.0 & 96.1 & 96.2 & 96.8\\ 
    \midrule
    Scenes-daytime &  85 & 87 & \textbf{91} & 92 & 92 & 95 & 95\\
    \midrule
    Handbags-color & 96.3 & \textbf{99.1} & 99.0 & 99.3 & 98.3 & 98.9 & 98.4\\
    Handbags-texture & 64.2 & 65.2 & 66.4 & \textbf{87.0} & 91.3 & 92.8 & 95.4\\
    \bottomrule
    \end{tabular}}%\vspace{-3mm}
    \caption{\small Classifier accuracy for different $D$ values. Boldface indicates the selected value for the semantic constraint. \vspace{-10mm}} 
    \label{tab:sem_const}
\end{table}

% \vspace{-1mm}
\minisection{Qualitative evaluation.}
Fig.~\ref{fig:morph_images}a and b show examples of young female and old female, respectively, and their corresponding translations to the other domain (old and young). 
As we can observe, the translations are realistic in general. 
DRIT tends to output uni-modal samples / generate only one distribution mode, while the other two methods also generate rich variations, including skin tones, hair color, beard/moustache variations, etc. 
However, MUNIT tends to generate diversity that includes changes in ethnicity and gender. 

In the case of the young female, gender is almost always changed due to the extreme bias towards males.
UDIT, on the other hand, preserves the wanted semantic properties and outputs diversity without unwanted changes.

\vspace{-1mm}
\minisection{Robustness to unwanted changes.} Here we evaluate how the identity constraint impacts gender and ethnicity changes compared to MUNIT and DRIT. 
Fig.~\ref{fig:morph_scores} shows the misclassification rate and drop in confidence of two classifiers, gender and ethnicity, trained on a disjoint subset of MORPH not used for translation.
We restrict our analysis to African and European, due to the very limited data in the other two ethnicities. 
The results show a drop in misclassification rate and a lower confidence drop when using UDIT, which are effective to alleviate gender bias (especially in females) and ethnicity bias (especially in Europeans). 
We also show ID distance, which achieves lower values for UDIT, indicating that identity is also better preserved. 
These results are in line with the observations in Fig.~\ref{fig:morph_images}.

\begin{figure*}[t]
    \centering 
    \begin{subfigure}[b]{\textwidth}
        \centering
        \includegraphics[width=\textwidth]{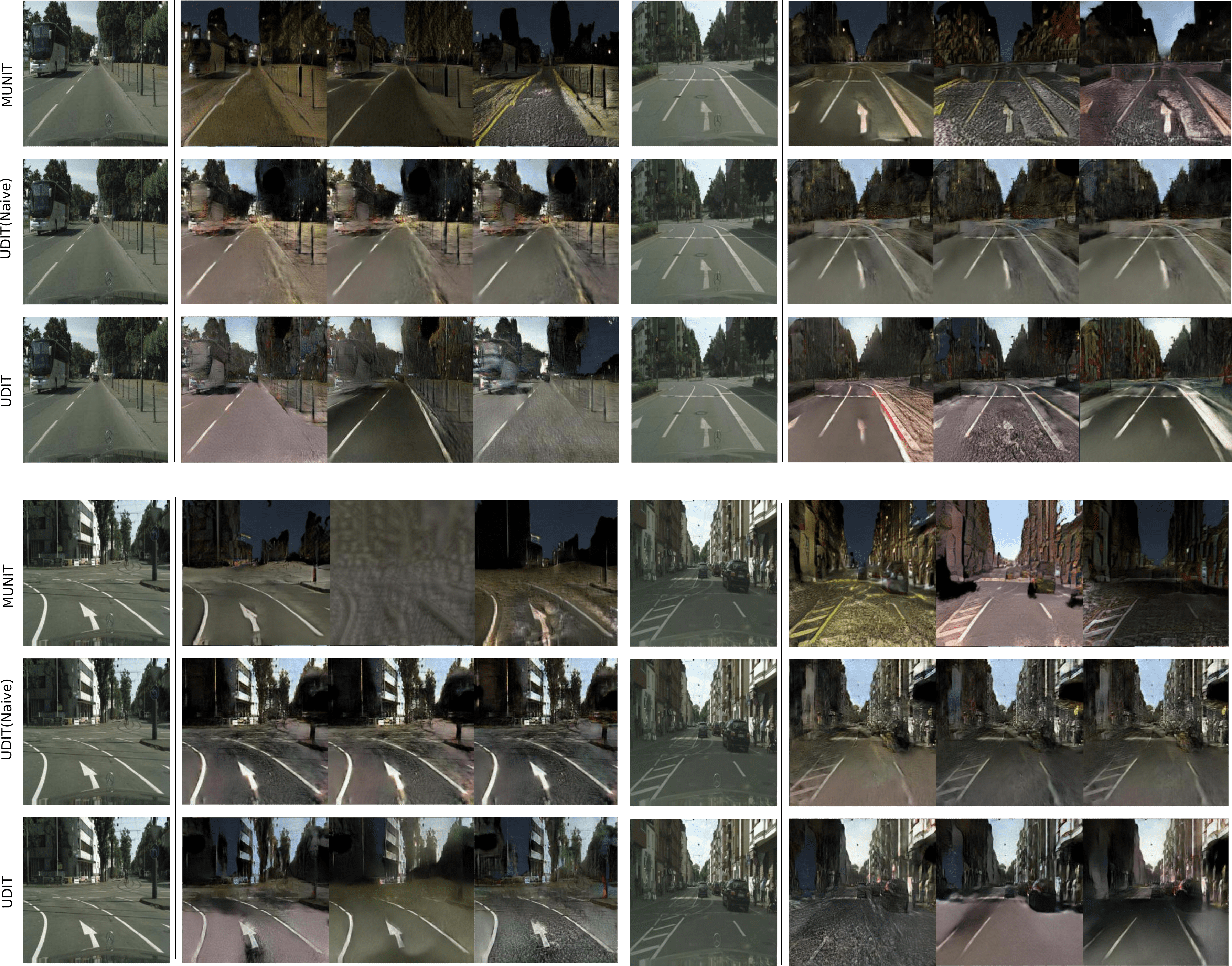}
        % \caption{\textit{flat} to \textit{textured} (input: red)}
    \end{subfigure}%
    % ~ 
    % \begin{subfigure}[b]{0.16\textwidth}
    %     \centering
    %       \includegraphics[width=\textwidth]{images/synthia_scores.pdf}
    %     % \caption{\textit{black} to \textit{red} (input: flat)}
    % \end{subfigure}
    % % \includegraphics[width=\textwidth]{images/synthia_qual.pdf}
    % \vspace{-3mm}
    \caption{\small Results on Cityscapes $\to$ Synthia-night. Example translations by MUNIT and UDIT with two variants of the semantic constraint. \vspace{-5mm}}
    \label{fig:synthia_results}
\end{figure*}

\vspace{-2mm}
\subsection{Cityscapes $\to$ Synthia-night}
% \vspace{-2mm}

\minisection{Semantic constraint.}
We train a binary classifier for daytime classification based on VGG16~\citep{simonyan2014very} using both real and synthetic images.
We use 6000 realistic images from BDD-100K~\citep{yu2018bdd100k} with a 50/50 daytime distribution.
As synthetic images we use 6000 images from a disjoint subset of Synthia~\citep{ros2016synthia}, also with a balanced class distribution.
We consider two semantic constraints.
The \emph{naive} variant employs features of the last convolutional layer, which have dimension $8\times 8 \times 512$.
Given the high dimensionality of these semantic features, the undesired information contained in them could potentially limit the model's translation ability or the output diversity.
For this reason, we also employ the \emph{reduced} semantic constraint variant presented in sec.~\ref{sec:semantic_extractor}, whose channel dimensions are reduced to $D$ by an additional $1\times1\times D$ layer.
In order to select a suitable dimensionality we train several classifiers with different $D$ values (Table~\ref{tab:sem_const}).
We select $D=16$ as it offers a good trade-off between small size and accuracy.

\minisection{Results.}
Figs.~\ref{fig:synthia_results} and ~\ref{fig:cityscape_scores}  present qualitative results and robustness measures respectively.
MUNIT translations mostly depict night scenes, as can be confirmed by the high misclassification rate and drop in confidence.
UDIT with naive constraint improves on this by preserving in the translations the input day-time.
However, the outputs have clearly limited diversity and lower quality.
UDIT with the reduced constraint achieves the overall best translations, both in terms of quality and wanted diversity.
This leads to remarkably low values on both robustness measures.

% \begin{figure}[t]
%     \centering
%          \includegraphics[width=\columnwidth]{images/handbags_both_qual.pdf}
%     \vspace{-6mm}
%     \caption{\small Example translations for Handbags-texture (left) and Handbags-color (right). Better viewed electronically, zoom might be necessary to appreciate the changes in texture.}
%     \label{fig:handbags_images}
% \end{figure}

\vspace{-2mm}
\subsection{Biased handbags} 
\begin{figure*}[t]
    \centering
    \includegraphics[width=0.9\textwidth]{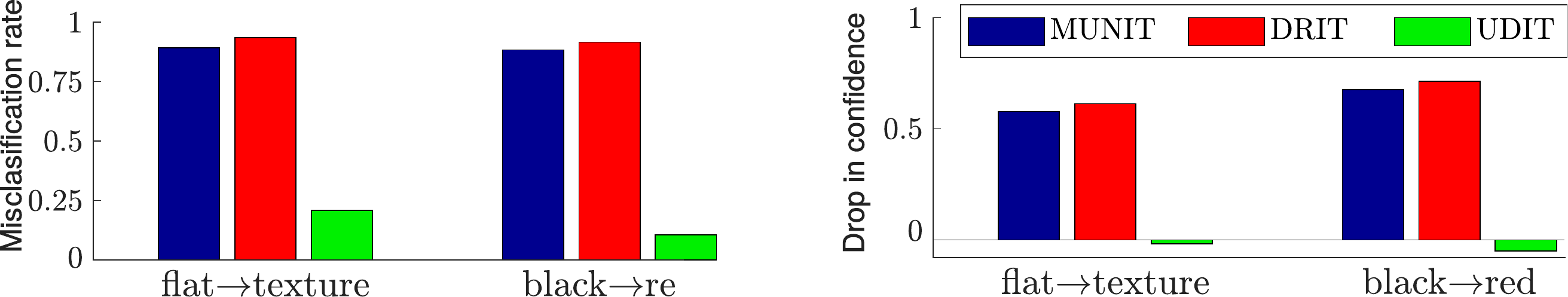}
    \vspace{-2mm}
    \caption{\small Robustness to bias on Biased handbags.\vspace{-4mm}}
    \label{fig:handbags_scores}
\end{figure*}

\begin{figure*}[t]
    \centering
        \includegraphics[width=0.9\textwidth]{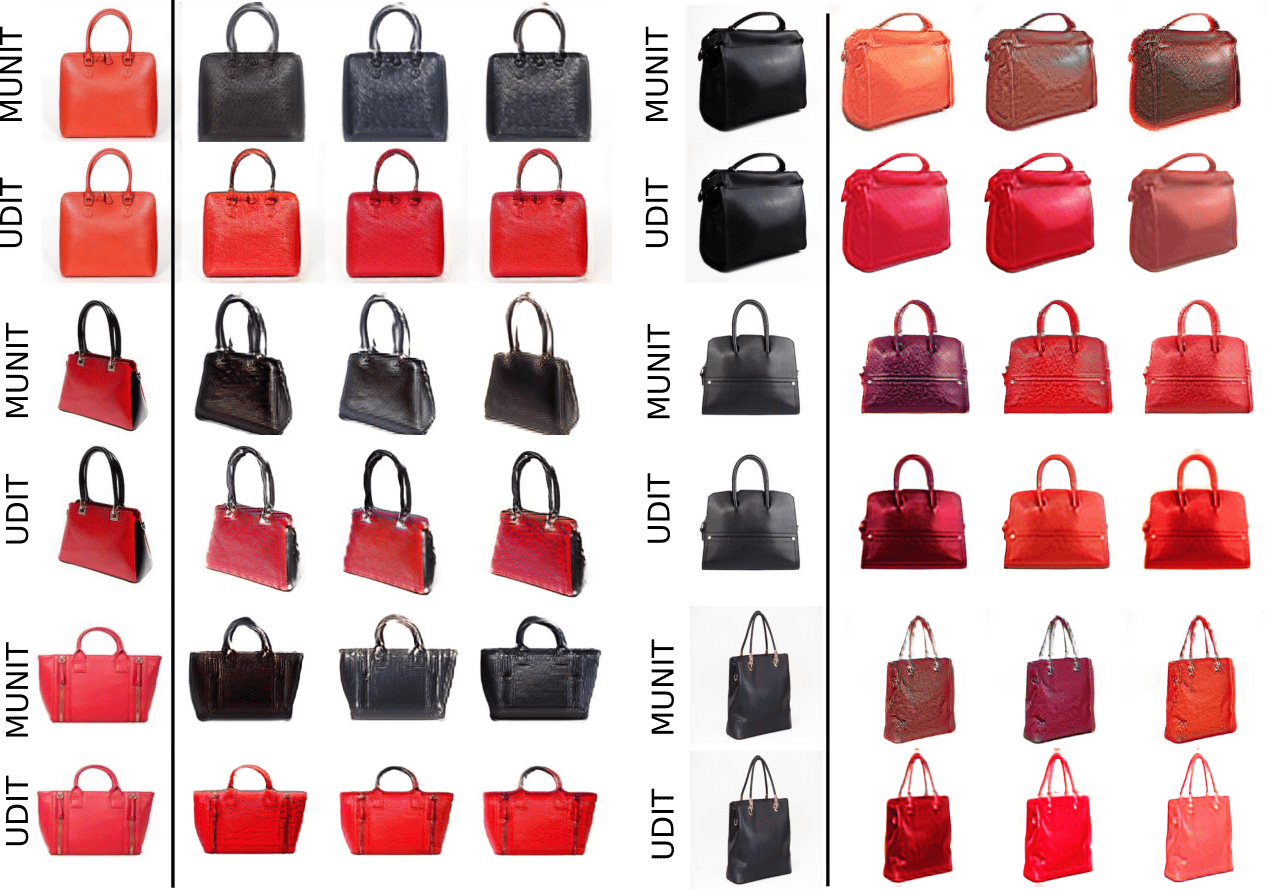}
    \caption{\small Example translations for Handbags-texture (left) and Handbags-color (right). Better viewed electronically, zoom might be necessary to appreciate the changes in texture.}\vspace{-6mm}
    \label{fig:handbags_images}
\end{figure*}

\minisection{Semantic constraint.}
We consider two different semantic constraints depending on the experiment. 
For Handbags-texture we train a color classifier selecting 500 images per color from~\citep{yu2018weakly}.
For Handbags-color, we gather images from the web searching for e.g.\ ``textured red handbag'' and verifying the downloaded images.
We use 1000 flat and 1000 textured handbags to train the classifier.
We only consider here the reduced variant of the semantic constraint.
Table~\ref{tab:sem_const} shows the accuracy results for the different $D$ values.
We select $D=8$ for color and $D=32$ for texture.
The overall lower accuracy of the texture classifier indicates that this is indeed a more subtle attribute, which in turn makes its recognition more challenging and increases the required dimensionality on the semantic features.

\minisection{Results.}
Fig.~\ref{fig:handbags_images} shows example results for these two experiments, evidencing how MUNIT succumbs to both types of biases.
UDIT, on the other hand, manages to perform the desired translation without introducing unwanted changes.
In general, the effects are more obvious for the color attribute as texture changes are harder to perceive.
We confirm the benefits of UDIT quantitatively in Fig.~\ref{fig:handbags_scores}.
MUNIT and DRIT present a notably high misclassification rate and drop in confidence for both experiments.
UDIT, instead, significantly increases the robustness to biases using a properly designed semantic constraint.

\vspace{-2mm}
\section{Conclusion}
\vspace{-2mm}
In this paper we tackle the problem of learning image translation models from biased datasets, which leads to unwanted changes in the output images. 
In order to address tdirection of MORPH.his problem, we propose the use of semantic constraints, which can effectively alleviate the effects of biases.
A properly designed semantic constraint allows for wanted diversity in the translations while preserving the desired semantic properties of the input image.
We evaluated the effectiveness of our UDIT model on faces, objects, and scenes.

\bibliographystyle{model2-names}
\bibliography{longstrings,refs}

\begin{thebibliography}{51}
\expandafter\ifx\csname natexlab\endcsname\relax\def\natexlab#1{#1}\fi
\providecommand{\url}[1]{\texttt{#1}}
\providecommand{\href}[2]{#2}
\providecommand{\path}[1]{#1}
\providecommand{\DOIprefix}{doi:}
\providecommand{\ArXivprefix}{arXiv:}
\providecommand{\URLprefix}{URL: }
\providecommand{\Pubmedprefix}{pmid:}
\providecommand{\doi}[1]{\href{http://dx.doi.org/#1}{\path{#1}}}
\providecommand{\Pubmed}[1]{\href{pmid:#1}{\path{#1}}}
\providecommand{\bibinfo}[2]{#2}
\ifx\xfnm\relax \def\xfnm[#1]{\unskip,\space#1}\fi
%Type = Inproceedings
\bibitem[{Almahairi et~al.(2018)Almahairi, Rajeswar, Sordoni, Bachman and
  Courville}]{almahairi2018augmented}
\bibinfo{author}{Almahairi, A.}, \bibinfo{author}{Rajeswar, S.},
  \bibinfo{author}{Sordoni, A.}, \bibinfo{author}{Bachman, P.},
  \bibinfo{author}{Courville, A.}, \bibinfo{year}{2018}.
\newblock \bibinfo{title}{Augmented cyclegan: Learning many-to-many mappings
  from unpaired data}, in: \bibinfo{booktitle}{International Conference on
  Machine Learning}.
%Type = Article
\bibitem[{Badrinarayanan et~al.(2017)Badrinarayanan, Kendall and
  Cipolla}]{badrinarayanan2017segnet}
\bibinfo{author}{Badrinarayanan, V.}, \bibinfo{author}{Kendall, A.},
  \bibinfo{author}{Cipolla, R.}, \bibinfo{year}{2017}.
\newblock \bibinfo{title}{Segnet: A deep convolutional encoder-decoder
  architecture for image segmentation}.
\newblock \bibinfo{journal}{{IEEE} Transactions on Pattern Analysis and Machine
  Intelligence} .
%Type = Article
\bibitem[{Bengio et~al.(2013)Bengio, Courville and
  Vincent}]{bengio2013representation}
\bibinfo{author}{Bengio, Y.}, \bibinfo{author}{Courville, A.},
  \bibinfo{author}{Vincent, P.}, \bibinfo{year}{2013}.
\newblock \bibinfo{title}{Representation learning: A review and new
  perspectives}.
\newblock \bibinfo{journal}{{IEEE} Transactions on Pattern Analysis and Machine
  Intelligence} .
%Type = Article
\bibitem[{Borji(2019)}]{borji2019pros}
\bibinfo{author}{Borji, A.}, \bibinfo{year}{2019}.
\newblock \bibinfo{title}{Pros and cons of gan evaluation measures}.
\newblock \bibinfo{journal}{Computer Vision and Image Understanding}
  \bibinfo{volume}{179}, \bibinfo{pages}{41--65}.
%Type = Inproceedings
\bibitem[{Bousmalis et~al.(2017)Bousmalis, Silberman, Dohan, Erhan and
  Krishnan}]{bousmalis2017unsupervised}
\bibinfo{author}{Bousmalis, K.}, \bibinfo{author}{Silberman, N.},
  \bibinfo{author}{Dohan, D.}, \bibinfo{author}{Erhan, D.},
  \bibinfo{author}{Krishnan, D.}, \bibinfo{year}{2017}.
\newblock \bibinfo{title}{Unsupervised pixel-level domain adaptation with
  generative adversarial networks}, in: \bibinfo{booktitle}{Proceedings of the
  IEEE Conference on Computer Vision and Pattern Recognition}.
%Type = Inproceedings
\bibitem[{Bousmalis et~al.(2016)Bousmalis, Trigeorgis, Silberman, Krishnan and
  Erhan}]{bousmalis2016domain}
\bibinfo{author}{Bousmalis, K.}, \bibinfo{author}{Trigeorgis, G.},
  \bibinfo{author}{Silberman, N.}, \bibinfo{author}{Krishnan, D.},
  \bibinfo{author}{Erhan, D.}, \bibinfo{year}{2016}.
\newblock \bibinfo{title}{Domain separation networks}, in:
  \bibinfo{booktitle}{Advances in Neural Information Processing Systems}.
%Type = Article
\bibitem[{Bozorgtabar et~al.(2019)Bozorgtabar, Rad, Ekenel and
  Thiran}]{bozorgtabar2019learn}
\bibinfo{author}{Bozorgtabar, B.}, \bibinfo{author}{Rad, M.S.},
  \bibinfo{author}{Ekenel, H.K.}, \bibinfo{author}{Thiran, J.P.},
  \bibinfo{year}{2019}.
\newblock \bibinfo{title}{Learn to synthesize and synthesize to learn}.
\newblock \bibinfo{journal}{Computer Vision and Image Understanding} .
%Type = Inproceedings
\bibitem[{Buolamwini and Gebru(2018)}]{buolamwini2018gender}
\bibinfo{author}{Buolamwini, J.}, \bibinfo{author}{Gebru, T.},
  \bibinfo{year}{2018}.
\newblock \bibinfo{title}{Gender shades: Intersectional accuracy disparities in
  commercial gender classification}, in: \bibinfo{booktitle}{Conference on
  Fairness, Accountability and Transparency}, pp. \bibinfo{pages}{77--91}.
%Type = Inproceedings
\bibitem[{Chen et~al.(2016)Chen, Duan, Houthooft, Schulman, Sutskever and
  Abbeel}]{chen2016infogan}
\bibinfo{author}{Chen, X.}, \bibinfo{author}{Duan, Y.},
  \bibinfo{author}{Houthooft, R.}, \bibinfo{author}{Schulman, J.},
  \bibinfo{author}{Sutskever, I.}, \bibinfo{author}{Abbeel, P.},
  \bibinfo{year}{2016}.
\newblock \bibinfo{title}{Infogan: Interpretable representation learning by
  information maximizing generative adversarial nets}, in:
  \bibinfo{booktitle}{Advances in Neural Information Processing Systems}.
%Type = Inproceedings
\bibitem[{Cordts et~al.(2016)Cordts, Omran, Ramos, Rehfeld, Enzweiler,
  Benenson, Franke, Roth and Schiele}]{Cordts2016Cityscapes}
\bibinfo{author}{Cordts, M.}, \bibinfo{author}{Omran, M.},
  \bibinfo{author}{Ramos, S.}, \bibinfo{author}{Rehfeld, T.},
  \bibinfo{author}{Enzweiler, M.}, \bibinfo{author}{Benenson, R.},
  \bibinfo{author}{Franke, U.}, \bibinfo{author}{Roth, S.},
  \bibinfo{author}{Schiele, B.}, \bibinfo{year}{2016}.
\newblock \bibinfo{title}{The cityscapes dataset for semantic urban scene
  understanding}, in: \bibinfo{booktitle}{Proceedings of the IEEE Conference on
  Computer Vision and Pattern Recognition}.
%Type = Article
\bibitem[{Daum{\'e}~III(2007)}]{daume2009frustratingly}
\bibinfo{author}{Daum{\'e}~III, H.}, \bibinfo{year}{2007}.
\newblock \bibinfo{title}{Frustratingly easy domain adaptation}.
\newblock \bibinfo{journal}{Proceedings of the Annual Meeting of the
  Association of Computational Linguistics} .
%Type = Inproceedings
\bibitem[{Fang et~al.(2013)Fang, Xu and Rockmore}]{fang2013unbiased}
\bibinfo{author}{Fang, C.}, \bibinfo{author}{Xu, Y.},
  \bibinfo{author}{Rockmore, D.N.}, \bibinfo{year}{2013}.
\newblock \bibinfo{title}{Unbiased metric learning: On the utilization of
  multiple datasets and web images for softening bias}, in:
  \bibinfo{booktitle}{Proceedings of the International Conference on Computer
  Vision}, pp. \bibinfo{pages}{1657--1664}.
%Type = Inproceedings
\bibitem[{Ganin and Lempitsky(2015)}]{ganin2015unsupervised}
\bibinfo{author}{Ganin, Y.}, \bibinfo{author}{Lempitsky, V.},
  \bibinfo{year}{2015}.
\newblock \bibinfo{title}{Unsupervised domain adaptation by backpropagation},
  in: \bibinfo{booktitle}{International Conference on Machine Learning}.
%Type = Inproceedings
\bibitem[{Gonzalez-Garcia et~al.(2018)Gonzalez-Garcia, van~de Weijer and
  Bengio}]{gonzalez2018image}
\bibinfo{author}{Gonzalez-Garcia, A.}, \bibinfo{author}{van~de Weijer, J.},
  \bibinfo{author}{Bengio, Y.}, \bibinfo{year}{2018}.
\newblock \bibinfo{title}{Image-to-image translation for cross-domain
  disentanglement}, in: \bibinfo{booktitle}{Advances in Neural Information
  Processing Systems}.
%Type = Inproceedings
\bibitem[{Goodfellow et~al.(2014)Goodfellow, Pouget-Abadie, Mirza, Xu,
  Warde-Farley, Ozair, Courville and Bengio}]{goodfellow2014generative}
\bibinfo{author}{Goodfellow, I.}, \bibinfo{author}{Pouget-Abadie, J.},
  \bibinfo{author}{Mirza, M.}, \bibinfo{author}{Xu, B.},
  \bibinfo{author}{Warde-Farley, D.}, \bibinfo{author}{Ozair, S.},
  \bibinfo{author}{Courville, A.}, \bibinfo{author}{Bengio, Y.},
  \bibinfo{year}{2014}.
\newblock \bibinfo{title}{Generative adversarial nets}, in:
  \bibinfo{booktitle}{Advances in Neural Information Processing Systems}.
%Type = Inproceedings
\bibitem[{Hendricks et~al.(2018)Hendricks, Burns, Saenko, Darrell and
  Rohrbach}]{hendricks2018women}
\bibinfo{author}{Hendricks, L.A.}, \bibinfo{author}{Burns, K.},
  \bibinfo{author}{Saenko, K.}, \bibinfo{author}{Darrell, T.},
  \bibinfo{author}{Rohrbach, A.}, \bibinfo{year}{2018}.
\newblock \bibinfo{title}{Women also snowboard: Overcoming bias in captioning
  models}, in: \bibinfo{booktitle}{Proceedings of the European Conference on
  Computer Vision}, \bibinfo{publisher}{Springer}. pp.
  \bibinfo{pages}{793--811}.
%Type = Inproceedings
\bibitem[{Herranz et~al.(2016)Herranz, Jiang and Li}]{herranz2016scene}
\bibinfo{author}{Herranz, L.}, \bibinfo{author}{Jiang, S.},
  \bibinfo{author}{Li, X.}, \bibinfo{year}{2016}.
\newblock \bibinfo{title}{Scene recognition with cnns: objects, scales and
  dataset bias}, in: \bibinfo{booktitle}{Proceedings of the IEEE Conference on
  Computer Vision and Pattern Recognition}, pp. \bibinfo{pages}{571--579}.
%Type = Inproceedings
\bibitem[{Howard et~al.(2017)Howard, Zhang and Horvitz}]{howard2017addressing}
\bibinfo{author}{Howard, A.}, \bibinfo{author}{Zhang, C.},
  \bibinfo{author}{Horvitz, E.}, \bibinfo{year}{2017}.
\newblock \bibinfo{title}{Addressing bias in machine learning algorithms: A
  pilot study on emotion recognition for intelligent systems}, in:
  \bibinfo{booktitle}{2017 IEEE Workshop on Advanced Robotics and its Social
  Impacts (ARSO)}, \bibinfo{organization}{IEEE}. pp. \bibinfo{pages}{1--7}.
%Type = Article
\bibitem[{Huang et~al.(2018)Huang, Liu, Belongie and
  Kautz}]{huang2018multimodal}
\bibinfo{author}{Huang, X.}, \bibinfo{author}{Liu, M.Y.},
  \bibinfo{author}{Belongie, S.}, \bibinfo{author}{Kautz, J.},
  \bibinfo{year}{2018}.
\newblock \bibinfo{title}{Multimodal unsupervised image-to-image translation}.
\newblock \bibinfo{journal}{Proceedings of the European Conference on Computer
  Vision} .
%Type = Inproceedings
\bibitem[{Isola et~al.(2017)Isola, Zhu, Zhou and Efros}]{isola2017pix2pix}
\bibinfo{author}{Isola, P.}, \bibinfo{author}{Zhu, J.Y.},
  \bibinfo{author}{Zhou, T.}, \bibinfo{author}{Efros, A.A.},
  \bibinfo{year}{2017}.
\newblock \bibinfo{title}{Image-to-image translation with conditional
  adversarial networks}, in: \bibinfo{booktitle}{Proceedings of the IEEE
  Conference on Computer Vision and Pattern Recognition}.
%Type = Article
\bibitem[{Jiang and Nachum(2019)}]{jiang2019identifying}
\bibinfo{author}{Jiang, H.}, \bibinfo{author}{Nachum, O.},
  \bibinfo{year}{2019}.
\newblock \bibinfo{title}{Identifying and correcting label bias in machine
  learning}.
\newblock \bibinfo{journal}{arXiv preprint arXiv:1901.04966} .
%Type = Inproceedings
\bibitem[{Khosla et~al.(2012)Khosla, Zhou, Malisiewicz, Efros and
  Torralba}]{khosla2012undoing}
\bibinfo{author}{Khosla, A.}, \bibinfo{author}{Zhou, T.},
  \bibinfo{author}{Malisiewicz, T.}, \bibinfo{author}{Efros, A.A.},
  \bibinfo{author}{Torralba, A.}, \bibinfo{year}{2012}.
\newblock \bibinfo{title}{Undoing the damage of dataset bias}, in:
  \bibinfo{booktitle}{Proceedings of the European Conference on Computer
  Vision}, \bibinfo{organization}{Springer}. pp. \bibinfo{pages}{158--171}.
%Type = Article
\bibitem[{Kim et~al.(2017)Kim, Cha, Kim, Lee and Kim}]{kim2017learning}
\bibinfo{author}{Kim, T.}, \bibinfo{author}{Cha, M.}, \bibinfo{author}{Kim,
  H.}, \bibinfo{author}{Lee, J.K.}, \bibinfo{author}{Kim, J.},
  \bibinfo{year}{2017}.
\newblock \bibinfo{title}{Learning to discover cross-domain relations with
  generative adversarial networks}.
\newblock \bibinfo{journal}{International Conference on Machine Learning} .
%Type = Article
\bibitem[{Lee et~al.(2018)Lee, Tseng, Huang, Singh and Yang}]{lee2018diverse}
\bibinfo{author}{Lee, H.Y.}, \bibinfo{author}{Tseng, H.Y.},
  \bibinfo{author}{Huang, J.B.}, \bibinfo{author}{Singh, M.},
  \bibinfo{author}{Yang, M.H.}, \bibinfo{year}{2018}.
\newblock \bibinfo{title}{Diverse image-to-image translation via disentangled
  representations}.
\newblock \bibinfo{journal}{Proceedings of the European Conference on Computer
  Vision} .
%Type = Article
\bibitem[{Lekic and Babic(2019)}]{CMGGANs}
\bibinfo{author}{Lekic, V.}, \bibinfo{author}{Babic, Z.}, \bibinfo{year}{2019}.
\newblock \bibinfo{title}{Automotive radar and camera fusion using generative
  adversarial networks}.
\newblock \bibinfo{journal}{Computer Vision and Image Understanding}
  \DOIprefix\doi{10.1016/j.cviu.2019.04.002}.
%Type = Inproceedings
\bibitem[{Levi and Hassner(2015)}]{levi2015age}
\bibinfo{author}{Levi, G.}, \bibinfo{author}{Hassner, T.},
  \bibinfo{year}{2015}.
\newblock \bibinfo{title}{Age and gender classification using convolutional
  neural networks}, in: \bibinfo{booktitle}{Proceedings of the IEEE Conference
  on Computer Vision and Pattern Recognition Workshops}, pp.
  \bibinfo{pages}{34--42}.
%Type = Inproceedings
\bibitem[{Liu et~al.(2017)Liu, Breuel and Kautz}]{liu2017unsupervised}
\bibinfo{author}{Liu, M.Y.}, \bibinfo{author}{Breuel, T.},
  \bibinfo{author}{Kautz, J.}, \bibinfo{year}{2017}.
\newblock \bibinfo{title}{Unsupervised image-to-image translation networks},
  in: \bibinfo{booktitle}{Advances in Neural Information Processing Systems},
  pp. \bibinfo{pages}{700--708}.
%Type = Article
\bibitem[{Liu et~al.(2019)Liu, Van De~Weijer and Bagdanov}]{liu2019exploiting}
\bibinfo{author}{Liu, X.}, \bibinfo{author}{Van De~Weijer, J.},
  \bibinfo{author}{Bagdanov, A.D.}, \bibinfo{year}{2019}.
\newblock \bibinfo{title}{Exploiting unlabeled data in cnns by self-supervised
  learning to rank}.
\newblock \bibinfo{journal}{{IEEE} Transactions on Pattern Analysis and Machine
  Intelligence} \bibinfo{volume}{41}, \bibinfo{pages}{1862--1878}.
%Type = Inproceedings
\bibitem[{Liu et~al.(2018)Liu, Yeh, Fu, Wang, Chiu and Wang}]{liu2018detach}
\bibinfo{author}{Liu, Y.C.}, \bibinfo{author}{Yeh, Y.Y.}, \bibinfo{author}{Fu,
  T.C.}, \bibinfo{author}{Wang, S.D.}, \bibinfo{author}{Chiu, W.C.},
  \bibinfo{author}{Wang, Y.C.F.}, \bibinfo{year}{2018}.
\newblock \bibinfo{title}{Detach and adapt: Learning cross-domain disentangled
  deep representation}, in: \bibinfo{booktitle}{Proceedings of the IEEE
  Conference on Computer Vision and Pattern Recognition}.
%Type = Inproceedings
\bibitem[{Mathieu et~al.(2016)Mathieu, Zhao, Zhao, Ramesh, Sprechmann and
  LeCun}]{mathieu2016disentangling}
\bibinfo{author}{Mathieu, M.F.}, \bibinfo{author}{Zhao, J.J.},
  \bibinfo{author}{Zhao, J.}, \bibinfo{author}{Ramesh, A.},
  \bibinfo{author}{Sprechmann, P.}, \bibinfo{author}{LeCun, Y.},
  \bibinfo{year}{2016}.
\newblock \bibinfo{title}{Disentangling factors of variation in deep
  representation using adversarial training}, in: \bibinfo{booktitle}{Advances
  in Neural Information Processing Systems}.
%Type = Inproceedings
\bibitem[{Parkhi et~al.(2015)Parkhi, Vedaldi and Zisserman}]{parkhi15deep}
\bibinfo{author}{Parkhi, O.M.}, \bibinfo{author}{Vedaldi, A.},
  \bibinfo{author}{Zisserman, A.}, \bibinfo{year}{2015}.
\newblock \bibinfo{title}{Deep face recognition}, in:
  \bibinfo{booktitle}{Proceedings of the British Machine Vision Conference}.
%Type = Article
\bibitem[{Patel et~al.(2015)Patel, Gopalan, Li and Chellappa}]{patel2015visual}
\bibinfo{author}{Patel, V.M.}, \bibinfo{author}{Gopalan, R.},
  \bibinfo{author}{Li, R.}, \bibinfo{author}{Chellappa, R.},
  \bibinfo{year}{2015}.
\newblock \bibinfo{title}{Visual domain adaptation: A survey of recent
  advances}.
\newblock \bibinfo{journal}{IEEE signal processing magazine}
  \bibinfo{volume}{32}, \bibinfo{pages}{53--69}.
%Type = Inproceedings
\bibitem[{Reed et~al.(2014)Reed, Sohn, Zhang and Lee}]{reed2014learning}
\bibinfo{author}{Reed, S.}, \bibinfo{author}{Sohn, K.}, \bibinfo{author}{Zhang,
  Y.}, \bibinfo{author}{Lee, H.}, \bibinfo{year}{2014}.
\newblock \bibinfo{title}{Learning to disentangle factors of variation with
  manifold interaction}, in: \bibinfo{booktitle}{International Conference on
  Machine Learning}.
%Type = Inproceedings
\bibitem[{Reed et~al.(2015)Reed, Zhang, Zhang and Lee}]{reed2015deep}
\bibinfo{author}{Reed, S.E.}, \bibinfo{author}{Zhang, Y.},
  \bibinfo{author}{Zhang, Y.}, \bibinfo{author}{Lee, H.}, \bibinfo{year}{2015}.
\newblock \bibinfo{title}{Deep visual analogy-making}, in:
  \bibinfo{booktitle}{Advances in Neural Information Processing Systems}.
%Type = Inproceedings
\bibitem[{Ricanek and Tesafaye(2006)}]{ricanek2006morph}
\bibinfo{author}{Ricanek, K.}, \bibinfo{author}{Tesafaye, T.},
  \bibinfo{year}{2006}.
\newblock \bibinfo{title}{Morph: A longitudinal image database of normal adult
  age-progression}, in: \bibinfo{booktitle}{Automatic Face and Gesture
  Recognition, 2006. FGR 2006. 7th International Conference on},
  \bibinfo{organization}{IEEE}. pp. \bibinfo{pages}{341--345}.
%Type = Inproceedings
\bibitem[{Ros et~al.(2016)Ros, Sellart, Materzynska, Vazquez and
  Lopez}]{ros2016synthia}
\bibinfo{author}{Ros, G.}, \bibinfo{author}{Sellart, L.},
  \bibinfo{author}{Materzynska, J.}, \bibinfo{author}{Vazquez, D.},
  \bibinfo{author}{Lopez, A.M.}, \bibinfo{year}{2016}.
\newblock \bibinfo{title}{The synthia dataset: A large collection of synthetic
  images for semantic segmentation of urban scenes}, in:
  \bibinfo{booktitle}{Proceedings of the IEEE Conference on Computer Vision and
  Pattern Recognition}, pp. \bibinfo{pages}{3234--3243}.
%Type = Article
\bibitem[{Russakovsky et~al.(2015)Russakovsky, Deng, Su, Krause, Satheesh, Ma,
  Huang, Karpathy, Khosla, Bernstein et~al.}]{russakovsky2015imagenet}
\bibinfo{author}{Russakovsky, O.}, \bibinfo{author}{Deng, J.},
  \bibinfo{author}{Su, H.}, \bibinfo{author}{Krause, J.},
  \bibinfo{author}{Satheesh, S.}, \bibinfo{author}{Ma, S.},
  \bibinfo{author}{Huang, Z.}, \bibinfo{author}{Karpathy, A.},
  \bibinfo{author}{Khosla, A.}, \bibinfo{author}{Bernstein, M.}, et~al.,
  \bibinfo{year}{2015}.
\newblock \bibinfo{title}{Imagenet large scale visual recognition challenge}.
\newblock \bibinfo{journal}{International Journal of Computer Vision}
  \bibinfo{volume}{115}, \bibinfo{pages}{211--252}.
%Type = Article
\bibitem[{Simonyan and Zisserman(2014)}]{simonyan2014very}
\bibinfo{author}{Simonyan, K.}, \bibinfo{author}{Zisserman, A.},
  \bibinfo{year}{2014}.
\newblock \bibinfo{title}{Very deep convolutional networks for large-scale
  image recognition}.
\newblock \bibinfo{journal}{arXiv preprint arXiv:1409.1556} .
%Type = Inproceedings
\bibitem[{Taigman et~al.(2017)Taigman, Polyak and
  Wolf}]{taigman2017unsupervised}
\bibinfo{author}{Taigman, Y.}, \bibinfo{author}{Polyak, A.},
  \bibinfo{author}{Wolf, L.}, \bibinfo{year}{2017}.
\newblock \bibinfo{title}{Unsupervised cross-domain image generation}, in:
  \bibinfo{booktitle}{International Conference on Learning Representations}.
%Type = Inproceedings
\bibitem[{Taigman et~al.(2014)Taigman, Yang, Ranzato and
  Wolf}]{taigman2014deepface}
\bibinfo{author}{Taigman, Y.}, \bibinfo{author}{Yang, M.},
  \bibinfo{author}{Ranzato, M.}, \bibinfo{author}{Wolf, L.},
  \bibinfo{year}{2014}.
\newblock \bibinfo{title}{Deepface: Closing the gap to human-level performance
  in face verification}, in: \bibinfo{booktitle}{Proceedings of the IEEE
  Conference on Computer Vision and Pattern Recognition}, pp.
  \bibinfo{pages}{1701--1708}.
%Type = Inproceedings
\bibitem[{Torralba and Efros(2011)}]{torralba2011unbiased}
\bibinfo{author}{Torralba, A.}, \bibinfo{author}{Efros, A.A.},
  \bibinfo{year}{2011}.
\newblock \bibinfo{title}{Unbiased look at dataset bias}, in:
  \bibinfo{booktitle}{Proceedings of the IEEE Conference on Computer Vision and
  Pattern Recognition}, \bibinfo{organization}{IEEE}. pp.
  \bibinfo{pages}{1521--1528}.
%Type = Inproceedings
\bibitem[{Wang et~al.(2018)Wang, van~de Weijer and Herranz}]{wang2018mix}
\bibinfo{author}{Wang, Y.}, \bibinfo{author}{van~de Weijer, J.},
  \bibinfo{author}{Herranz, L.}, \bibinfo{year}{2018}.
\newblock \bibinfo{title}{Mix and match networks: encoder-decoder alignment for
  zero-pair image translation}, in: \bibinfo{booktitle}{Proceedings of the IEEE
  Conference on Computer Vision and Pattern Recognition}.
%Type = Inproceedings
\bibitem[{Yi et~al.(2017)Yi, Zhang, Tan and Gong}]{yi2017dualgan}
\bibinfo{author}{Yi, Z.}, \bibinfo{author}{Zhang, H.R.}, \bibinfo{author}{Tan,
  P.}, \bibinfo{author}{Gong, M.}, \bibinfo{year}{2017}.
\newblock \bibinfo{title}{Dualgan: Unsupervised dual learning for
  image-to-image translation.}, in: \bibinfo{booktitle}{Proceedings of the
  International Conference on Computer Vision}, pp.
  \bibinfo{pages}{2868--2876}.
%Type = Article
\bibitem[{Yu et~al.(2018a)Yu, Xian, Chen, Liu, Liao, Madhavan and
  Darrell}]{yu2018bdd100k}
\bibinfo{author}{Yu, F.}, \bibinfo{author}{Xian, W.}, \bibinfo{author}{Chen,
  Y.}, \bibinfo{author}{Liu, F.}, \bibinfo{author}{Liao, M.},
  \bibinfo{author}{Madhavan, V.}, \bibinfo{author}{Darrell, T.},
  \bibinfo{year}{2018}a.
\newblock \bibinfo{title}{Bdd100k: A diverse driving video database with
  scalable annotation tooling}.
\newblock \bibinfo{journal}{Proceedings of the European Conference on Computer
  Vision} .
%Type = Inproceedings
\bibitem[{Yu et~al.(2018b)Yu, Cheng and van~de Weijer}]{yu2018weakly}
\bibinfo{author}{Yu, L.}, \bibinfo{author}{Cheng, Y.}, \bibinfo{author}{van~de
  Weijer, J.}, \bibinfo{year}{2018}b.
\newblock \bibinfo{title}{Weakly supervised domain-specific color naming based
  on attention}, in: \bibinfo{booktitle}{Proceedings of the International
  Conference on Pattern Recognition}, \bibinfo{organization}{IEEE}. pp.
  \bibinfo{pages}{3019--3024}.
%Type = Article
\bibitem[{Zhang et~al.(2018a)Zhang, Gonzalez-Garcia, van~de Weijer, Danelljan
  and Khan}]{zhang2018synthetic}
\bibinfo{author}{Zhang, L.}, \bibinfo{author}{Gonzalez-Garcia, A.},
  \bibinfo{author}{van~de Weijer, J.}, \bibinfo{author}{Danelljan, M.},
  \bibinfo{author}{Khan, F.S.}, \bibinfo{year}{2018}a.
\newblock \bibinfo{title}{Synthetic data generation for end-to-end thermal
  infrared tracking}.
\newblock \bibinfo{journal}{{IEEE} Transactions on Image Processing}
  \bibinfo{volume}{28}, \bibinfo{pages}{1837--1850}.
%Type = Inproceedings
\bibitem[{Zhang et~al.(2018b)Zhang, Isola, Efros, Shechtman and
  Wang}]{zhang2018perceptual}
\bibinfo{author}{Zhang, R.}, \bibinfo{author}{Isola, P.},
  \bibinfo{author}{Efros, A.A.}, \bibinfo{author}{Shechtman, E.},
  \bibinfo{author}{Wang, O.}, \bibinfo{year}{2018}b.
\newblock \bibinfo{title}{The unreasonable effectiveness of deep networks as a
  perceptual metric}, in: \bibinfo{booktitle}{Proceedings of the IEEE
  Conference on Computer Vision and Pattern Recognition}.
%Type = Inproceedings
\bibitem[{Zhao et~al.(2018)Zhao, Ren, Yuan, Song, Goodman and
  Ermon}]{zhao2018bias}
\bibinfo{author}{Zhao, S.}, \bibinfo{author}{Ren, H.}, \bibinfo{author}{Yuan,
  A.}, \bibinfo{author}{Song, J.}, \bibinfo{author}{Goodman, N.},
  \bibinfo{author}{Ermon, S.}, \bibinfo{year}{2018}.
\newblock \bibinfo{title}{Bias and generalization in deep generative models: An
  empirical study}, in: \bibinfo{booktitle}{Advances in Neural Information
  Processing Systems}, pp. \bibinfo{pages}{10815--10824}.
%Type = Inproceedings
\bibitem[{Zhu et~al.(2017a)Zhu, Park, Isola and Efros}]{zhu2017unpaired}
\bibinfo{author}{Zhu, J.Y.}, \bibinfo{author}{Park, T.},
  \bibinfo{author}{Isola, P.}, \bibinfo{author}{Efros, A.A.},
  \bibinfo{year}{2017}a.
\newblock \bibinfo{title}{Unpaired image-to-image translation using
  cycle-consistent adversarial networks}, in: \bibinfo{booktitle}{Proceedings
  of the IEEE Conference on Computer Vision and Pattern Recognition}.
%Type = Inproceedings
\bibitem[{Zhu et~al.(2017b)Zhu, Zhang, Pathak, Darrell, Efros, Wang and
  Shechtman}]{zhu2017toward}
\bibinfo{author}{Zhu, J.Y.}, \bibinfo{author}{Zhang, R.},
  \bibinfo{author}{Pathak, D.}, \bibinfo{author}{Darrell, T.},
  \bibinfo{author}{Efros, A.A.}, \bibinfo{author}{Wang, O.},
  \bibinfo{author}{Shechtman, E.}, \bibinfo{year}{2017}b.
\newblock \bibinfo{title}{Toward multimodal image-to-image translation}, in:
  \bibinfo{booktitle}{Advances in Neural Information Processing Systems}, pp.
  \bibinfo{pages}{465--476}.
%Type = Misc
\bibitem[{Zou and Schiebinger(2018)}]{zou2018ai}
\bibinfo{author}{Zou, J.}, \bibinfo{author}{Schiebinger, L.},
  \bibinfo{year}{2018}.
\newblock \bibinfo{title}{Ai can be sexist and racist—it’s time to make it
  fair}.

\end{thebibliography}

\clearpage
\section*{Appendix}
\label{Appendix}
%\section{Network architecture}
 Tables~\ref{table:encoders of content enconder}-\ref{table:the image discriminator} show the architectures of the content encoder, style encoder, image decoder and discriminator used in the cross-modal experiment.  The used abbreviations are shown in Table~\ref{table:Abbreviation name}.
%image content encoder
\begin{table}[h]
\setlength{\arrayrulewidth}{0.6\arrayrulewidth}% 50% thicker
\centering
\resizebox{0.7\columnwidth}{!}{
\begin{tabular}{cc|ccc}
\hline
Layer &Input $\rightarrow $Output    &Kernel, stride, pad\\ 
\hline  
conv1   & [4,128, 128,3] $\rightarrow$ [4,128, 128, 64] & [7,7], 1, 3\\ 
IN1   & [4,128, 128, 64] $\rightarrow$ [4,128, 128, 64] &  -, -, -\\ 
pool1 (max) & [4,128, 128, 64] $\rightarrow$[4,64, 64, 64]+indices1 & [2,2], 2, - \\
\hline
conv2   & [4,64, 64,64] $\rightarrow$ [4,64, 64,128] & [7,7], 1, 3\\ 
IN2   & [4,64, 64,128] $\rightarrow$ [4,64, 64,128] &  -, -, -\\ 
pool2 (max) & [4,64, 64,128] $\rightarrow$[4,32, 32,128]+indices2 & [2,2], 2, - \\
\hline
conv3   & [4,32, 32,128] $\rightarrow$ [4,32, 32,256] & [7,7], 1, 3\\ 
IN3   &  [4,32, 32,256] $\rightarrow$ [4,32, 32,256] &  -, -, -\\ 
pool3 (max) &  [4,32, 32,256] $\rightarrow$ [4,16, 16,256]+indices3 & [2,2], 2, - \\
\hline
RB(IN)4-9   & [4,16, 16,256] $\rightarrow$ [4,16, 16,256] & [7,7], 1, 3\\ 

\hline
\end{tabular}
}
\caption{Content encoder.}
\label{table:encoders of content enconder}
\end{table}
%image style encoder 
\begin{table}[h]
\setlength{\arrayrulewidth}{1\arrayrulewidth}% 50% thicker
\centering
\resizebox{0.7\columnwidth}{!}{
\begin{tabular}{cc|ccc}
\hline
Layer &Input $\rightarrow $Output    &Kernel, stride, pad\\ 
\hline  
conv1   & [4,128, 128,3] $\rightarrow$ [4,128, 128, 64] & [7,7], 1, 3\\ 

relu1 & [4,128, 128, 64] $\rightarrow$[4,64, 64, 64] & -, -, - \\
\hline
conv2   & [4,64, 64,64] $\rightarrow$ [4,32, 32,128] & [4, 4], 2, 1\\ 

relu2 & [4,32, 32,128] $\rightarrow$[4,32, 32,128] & -, -, - \\
\hline
conv3   & [4,32, 32,128] $\rightarrow$ [4,16, 16,256] & [4,4], 2, 1\\ 
 
relu3 &  [4,16, 16,256] $\rightarrow$ [4,16, 16,256] & -, -, - \\
\hline
GAP   & [4,16, 16,256] $\rightarrow$ [4,1, 1,256] & -, -,-\\ 
\hline
conv4  & [4,1, 1,256] $\rightarrow$ [4,1, 1,8] & [1, 1],1,0\\ 
\hline
\end{tabular}
}
\caption{Style encoder.}
\label{table:encoders of style enconder}
\end{table}
%\bibliographystyle{ieee}
%\bibliography{shortstrings,egbib}

%image decoder

\begin{table}[!htb]
    \centering
    \begin{subtable}{.4\linewidth}
      \centering
          \resizebox{0.9\columnwidth}{!}{
            \begin{tabular}{ll}
                \hline
                Layer &Input $\rightarrow $Output    \\ 
                \hline  
                linear1   & [4, 8] $\rightarrow$ [4, 256] \\ 
                relu1 & [4, 256] $\rightarrow$[4, 256] \\
                \hline
                linear2   & [4, 256] $\rightarrow$ [4, 256] \\ 
                relu2 & [4, 256] $\rightarrow$[4, 256] \\
                \hline
                linear3   & [4, 256] $\rightarrow$ [4, 256] \\ 
                reshape & [4, 256] $\rightarrow$[4,1,1, 256] \\
                \hline
            \end{tabular}
            }
     \caption{affine parameter $\mu$}
    \end{subtable}%
    \begin{subtable}{.4\linewidth}
      \centering
          \resizebox{0.9\columnwidth}{!}{
            \begin{tabular}{ll}
                \hline
                Layer &Input $\rightarrow $Output    \\ 
                \hline  
                linear1   & [4, 8] $\rightarrow$ [4, 256] \\ 
                relu1 & [4, 256] $\rightarrow$[4, 256] \\
                \hline
                linear2   & [4, 256] $\rightarrow$ [4, 256] \\ 
                relu2 & [4, 256] $\rightarrow$[4, 256] \\
                \hline
                linear3   & [4, 256] $\rightarrow$ [4, 256] \\ 
                reshape & [4, 256] $\rightarrow$[4,1,1, 256] \\
                \hline
            \end{tabular}
            }
        \caption{affine parameter $\sigma$}
    \end{subtable} 
    \caption{Networks for the estimation of the affine parameters that are used in the AdaIN layer. The parameters (a) $\mu$ and (b) $\sigma$ scale and shift the normalized content, respectively. Note that (a) and (b) share the first two layers.}
    \label{table:encoder for affine parameters}
\end{table}
%\clearpage
%image generator
\begin{table}[h]
\setlength{\arrayrulewidth}{1\arrayrulewidth}% 50% thicker
\centering
\resizebox{0.7\columnwidth}{!}{
\begin{tabular}{cc|ccc}
\hline
Layer &Input $\rightarrow $Output    &Kernel, stride, pad\\ 
\hline  
RB(AdaIN)1-6    &  ($\mu, \sigma$) +[4,16, 16,256] $\rightarrow$ [4,16, 16,256] & [7,7], 1, 3\\ 

\hline
unpool1   & indices3 + [4,16, 16,256] $\rightarrow$ [4,32, 32,256] & [2, 2], 2, -\\ 
conv1   & [4,32, 32,256] $\rightarrow$ [4,32, 32,128] & [7,7], 1, 3\\ 
IN1   & [4,32, 32,128] $\rightarrow$ [4,32, 32,128] &  -, -, -\\ 
\hline
unpool2   & indices2 + [4,32, 32,128] $\rightarrow$ [4, 64, 64,128] & [2, 2], 2, -\\ 
conv2   & [4, 64, 64,128] $\rightarrow$  [4, 64, 64,64] & [7,7], 1, 3\\ 
IN2   & [4, 64, 64,64]$\rightarrow$   [4, 64, 64,64] &  -, -, -\\ 
\hline
unpool3   & indices1 + [4, 64, 64,64] $\rightarrow$ [4, 128, 128,64] & [2, 2], 2, -\\ 
conv3   & [4, 128, 128,64] $\rightarrow$  [4, 128, 128,3] & [7,7], 1, 3\\ 
\hline
\end{tabular}
}
\caption{Decoder (Image generator).}
\label{table:the image generator}
\end{table}

%discriminator
\FloatBarrier
\begin{table}[h]
\centering
\setlength{\arrayrulewidth}{0.9\arrayrulewidth}% 50% thicker
\resizebox{0.7\columnwidth}{!}{
\begin{tabular}{cc|ccc}
\hline
Layer &Input $\rightarrow $Output    &Kernel, stride, pad\\ 
\hline  
conv1    &  [4,128, 128,3] $\rightarrow$ [4,64, 64,64] & [4,4], 2, 1\\ 
lrelu1    &  [4,64, 64,64] $\rightarrow$ [4,64, 64,64] & -, -, -\\
\hline
conv2    &  [4,64, 64,64] $\rightarrow$ [4,32, 32,128] & [4,4], 2, 1\\ 
lrelu2    &  [4,32, 32,128] $\rightarrow$ [4,32, 32,128] & -, -, -\\
\hline
conv3    &   [4,32, 32,128]  $\rightarrow$   [4,16, 16,256]  & [4,4], 2, 1\\ 
lrelu3    &  [4,16, 16,256] $\rightarrow$ [4,16, 16,256] & -, -, -\\
\hline
conv4    &   [4,16, 16,256]  $\rightarrow$   [4,8, 8,512]  & [4,4], 2, 1\\ 
lrelu4    &[4,8, 8,512]  $\rightarrow$[4,8, 8,512]  & -, -, -\\
\hline
conv5    &[4,8, 8,512]  $\rightarrow$[4,8, 8,1]  & [1,1], 1, 0\\ 
\hline

\end{tabular}
}
\caption{Architecture for the discrim Loss specificationinator for $128\times128$ input. The discriminators for $64\times64$, and $32\times32$ use the same convolutional architecture.}
\label{table:the image discriminator}
\end{table}

%image content encoder
\begin{table}[h]
\centering
\resizebox{0.7\columnwidth}{!}{
\begin{tabular}{cc}
\hline
Abbreviation & Name   \\ 

\hline  
pool    &  pooling layer \\ 
\hline
unpool    &  unpooling layer \\ 
\hline
lrelu    &  leaky relu layer \\ 
\hline
concat    &  concatenate layer \\ 
\hline  
conv    &  convolutional layer\\ 
\hline 
linear    &  fully connection layer \\ 
\hline
IN    &  instance normalization layer\\ 
\hline  
GAP    &  global average pooling layer \\ 
\hline
RB(IN)    &  residual block layer using instance normalization \\ 
\hline
RB(AdaIN)    &  residual block layer using adaptive instance normalization \\ 
\hline

\end{tabular}
}
        \caption{Abbreviations used in other tables.}
\label{table:Abbreviation name}

\end{table}

\end{document}